\pdfoutput=1
\documentclass[11pt]{article}

\usepackage{acl}

\usepackage{times}
\usepackage{latexsym}

\usepackage{listings}
\usepackage{enumitem}
\usepackage{multirow}

\usepackage[T1]{fontenc}
\usepackage[utf8]{inputenc}

\usepackage{microtype}

\title{Extreme Miscalibration and the Illusion of Adversarial Robustness}

\author{Vyas Raina*$^{\dagger}$ \hspace{1.0em} Samson Tan*$^{\ddag}$ \hspace{1.0em} Volkan Cevher$^{\ddag, \P}$ \\
        \textbf{Aditya Rawal}$^{\ddag}$\ \hspace{1.0em} \textbf{Sheng Zha}$^{\ddag}$ \hspace{1.0em} \textbf{George Karypis}$^{\ddag}$\\ \\
$^\dagger$University of Cambridge \quad $^\ddag$Amazon\\
$^\P$LIONS, IEM, STI, Ecole Polytechnique Federale de Lausanne\\
\texttt{vr313@cam.ac.uk\quad
samson@amazon.com}
}
\usepackage{amssymb}
\usepackage{amsmath}
\usepackage{bm}
\DeclareMathOperator*{\argmax}{arg\,max}
\DeclareMathOperator*{\argmin}{arg\,min}

\usepackage{booktabs}

\usepackage{graphicx}
\usepackage{caption}
\usepackage{subcaption}

\begin{document}
\maketitle

\footnotetext[0\def\thefootnote{}]{\noindent $^*$Equal contribution.
This paper reports on the work VR did whilst at Amazon Web Services. VC holds concurrent appointments as an Amazon Scholar and as a faculty at EPFL. This paper describes the work performed at Amazon.
}

\begin{abstract}
Deep learning-based Natural Language Processing (NLP) models are vulnerable to adversarial attacks, where small perturbations can cause a model to misclassify. Adversarial Training (AT) is often used to increase model robustness. However, we have discovered an intriguing phenomenon: deliberately or accidentally miscalibrating models masks gradients in a way that interferes with adversarial attack search methods, giving rise to an apparent increase in robustness. We show that this observed gain in robustness is an illusion of robustness (IOR), and demonstrate how an adversary can perform various forms of test-time temperature calibration to nullify the aforementioned interference and allow the adversarial attack to find adversarial examples. Hence, we urge the NLP community to incorporate test-time temperature scaling into their robustness evaluations to ensure that any observed gains are genuine. Finally, we show how the temperature can be scaled during \textit{training} to improve genuine robustness.
\end{abstract}

\section{Introduction}

% \begin{figure*}[htb!]
%      \centering
%      \begin{subfigure}[b]{0.32\linewidth}
%          \centering
%          \includegraphics[width=\columnwidth]{Figures/Concept/non-robust.png}
%          \caption{Non-robust}
%      \end{subfigure}
%      \hfill
%      \begin{subfigure}[b]{0.32\linewidth}
%          \centering
%          \includegraphics[width=1.04\columnwidth]{Figures/Concept/illusion.png}
%          \caption{Illusion of Robustness}
%      \end{subfigure}
%      \hfill
%      \begin{subfigure}[b]{0.32\linewidth}
%          \centering
%          \includegraphics[width=\columnwidth]{Figures/Concept/robust.png}
%          \caption{True Robustness}
%      \end{subfigure}
%         \caption{Conceptualization of model robustness}
%         \label{fig:concept}
% \end{figure*}

\begin{figure}[htb!]
    \centering
    \includegraphics[width=\columnwidth]{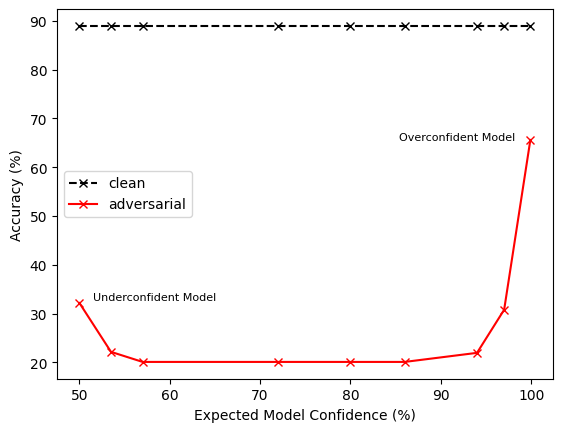}
    \caption{Accuracy on adversarial examples from out-of-the-box adversarial attack for models with different average predicted class confidence, $E_{p(\mathbf x)} [P_{\hat{\theta}}(\hat c|\mathbf x)]$. Extremely overconfident and underconfident models show increased robustness. We reveal that this increased robustness is merely an \textit{illusion of robustness}.}
    \label{fig:extreme}
\end{figure}

Deep learning Natural Language Processing (NLP) models are able to perform well in a range of tasks~\citep{manning-etal-2014-stanford}. However, these NLP models are susceptible to adversarial attacks, where perturbing clean input text samples slightly (accidentally or maliciously by an adversary) can lead to a NLP model misclassifying the perturbed input \citep{jia-liang-2017-adversarial}. However, the emergence of the Adversarial Training (AT) paradigm~\citep{goodfellow2015explaining} has shown some success in training models to be more robust to these small adversarial perturbations. Here, the traditional training process is adapted to minimize the empirical risk associated with a ``robustness loss'' as opposed to the risk associated with the standard loss for clean input samples. The robustness loss is the standard loss applied to the worst-case (loss maximizing) adversarial sample for each training sample. In NLP, due to the discrete nature of the text, this adversarial training min-max formulation is particularly challenging as the inner maximization is computationally expensive~\citep{DBLP:journals/corr/abs-2109-00544}. Nevertheless, a variety of approaches have been proposed in the robustness literature, ranging from augmentation of the training set with adversarial examples for a specific model, to sophisticated token-embedding space optimizations for the inner maximization step~\citep{wang-etal-2019-deep, goyal2023survey}.

Although many NLP AT methods appear effective in boosting model robustness, we argue that, in some cases, the observed improvement is merely an \textit{illusion of robustness} (IOR) that can be easily circumvented. In the computer vision literature, \citet{DBLP:journals/corr/abs-1802-00420} show that certain modeling decisions can give rise to such an illusion via gradient obfuscation \citep{10.1145/3052973.3053009}. We build on these findings by showing that IORs can also emerge easily in NLP AT due to highly miscalibrated models. Concretely, we:
\begin{itemize}[leftmargin=*]
    \item Argue that the extreme class confidence of miscalibrated models~\citep{DBLP:journals/corr/GuoPSW17} gives rise to an illusion of robustness by disrupting the adversarial attacks' search processes.
    \item Demonstrate this by intentionally creating highly over- and underconfident models at inference time, and by applying an existing adversarial training technique from the literature. These models appear to be up to three times more robust than the baseline. However, we reveal this robustness to be an illusion that largely evaporates when naive model calibration is applied.
    \item Further introduce a test-time adversarial temperature optimization algorithm that practically nullifies the perceived robustness gains and demonstrate its efficacy on a range of encoder models, classification datasets, and adversarial attacks.
     \item Finally use the above insights to improve \textit{true} robustness to unseen attacks by increasing the training temperature, demonstrating its efficacy alone and in combination with other AT methods. In contrast to \citet{DBLP:journals/corr/PapernotMWJS15}, we do not perform model distillation but only training-time temperature scaling as the adversarial defense.
\end{itemize}

In light of our findings, we urge the community to perform test-time temperature scaling during all robustness evaluations to ensure that the observed robustness is genuine and not merely an illusion. To our knowledge, we are the first to demonstrate this phenomenon in NLP models and propose 
training-time temperature scaling as an adversarial defense.

\section{Background and Related Work}

\subsection{Adversarial Attacks}\label{sec:attacks}

An untargeted adversarial attack is able to fool a classification system, $\mathcal F()$ with trained parameters $\hat{\theta}$, by perturbing an input sample, $\mathbf x$ to generate an adversarial example $\tilde{\mathbf x}$ to cause a change in the predicted class, 
\begin{equation}
    \mathcal F(\mathbf x; \hat{\theta})\neq\mathcal F(\mathbf{\tilde x}; \hat{\theta}).
\end{equation}
Traditional adversarial attack definitions \citep{szegedy2014intriguing} require the perturbation to be \textit{imperceptible} as per human perception. In NLP it can be challenging to measure imperceptibility. Following notation in \citet{raina2023sample} the distance between the original, clean sample and the adversarial example is limited as per a proxy distance measure $\mathcal{G}(\mathbf x, \mathbf{\tilde x}) \leq\epsilon$.

% Following \citet{morris2020textattack} and \citet{raina2023sample}, we can separate modern NLP imperceptibility constraints into two categories: 1) pre-transformation constraints, which limit the changes that can be made to a clean sample $\mathbf x$, such that an adversarial example is limited to a specific set of sequences $\mathbf{\tilde x}\in\mathcal A(\mathbf x)$; and 2) distance-based constraints, which aim to mathematically limit the distance between the original, clean sample and the adversarial example using a proxy distance measure $\mathcal{G}(\mathbf x, \mathbf{\tilde x}) \leq\epsilon$.

A plethora of adversarial attack approaches have been proposed for efficiently discovering adversarial examples for NLP models~\citep{DBLP:journals/corr/abs-1804-07998, DBLP:journals/corr/abs-1812-05271, DBLP:journals/corr/abs-1801-04354, DBLP:journals/corr/abs-1909-06723, ren-etal-2019-generating, DBLP:journals/corr/abs-1907-11932, tan-etal-2020-morphin,garg-ramakrishnan-2020-bae, DBLP:journals/corr/abs-2004-09984,tan-joty-2021-code}. Many of the popular attack approaches are implemented in the TextAttack library~\citep{morris2020textattack}. 

% These attack approaches can be classed as either whitebox attacks, where the adversary has full access to the model parameters or blackbox attacks, where the adversary can only access input-output pairs from the model~\cite{Tabassi2019ATA}.

\subsection{Adversarial Training} \label{sec:at}
Standard supervised training methods seek to find model parameters, $\hat{\theta}$, that minimizes the empirical risk (for a dataset of $\mathbf x\sim p(\mathbf x)$), characterised by a loss function,
\begin{equation}\label{eqn:standard}
    \hat{\theta} = \argmin_{\theta} \underset{\mathbf x \sim p(\mathbf x)}{\mathbb E}[\mathcal L(\mathbf x, \theta)].
\end{equation}
Adversarial Training (AT;~\citealp{goodfellow2015explaining}) adapts the objective to minimize the empirical risk associated with the \textit{worst-case} adversarial example, $\mathbf{\tilde x}$, such that we are minimizing a \textit{robust loss},
\begin{equation} \label{eqn:robust}
    \hat{\theta} = \argmin_{\theta} \underset{\mathbf x \sim p(\mathbf x)}{\mathbb E}\left [\max_{\substack{\mathbf{\tilde x}:\\ \mathcal{G}(\mathbf x, \mathbf{\tilde x}, ) \leq\epsilon,\hspace{0.2em} \mathbf{\tilde x}\in\mathcal A}} \mathcal L(\mathbf{\tilde x}, \theta)\right ].
\end{equation}
It is too computationally expensive to perform the inner maximization step to find textual adversarial examples in each step of training. A group of AT methods speed-up this optimization step by finding adversarial examples in the token embedding space, which allows for faster gradient-based approaches: PGD-K~\citep{madry2018towards}, FreeLB~\citep{Zhu2020FreeLB}, TA-VAT~\citep{DBLP:journals/corr/abs-2004-14543}, InfoBERT~\citep{DBLP:journals/corr/abs-2010-02329}. However, limited success of these approaches has been attributed to perturbations in the embedding space being unrepresentative of real textual adversarial attacks. Hence, AT methods such as Adversarial Sparse Convex Combination (ASCC;~\citealp{DBLP:journals/corr/abs-2107-13541}) and Dirichlet Neighborhood Ensemble (DNE;~\citealp{DBLP:journals/corr/abs-2006-11627}) identify a more sensible embedding perturbation space, which they define as the convex hull of word synonyms. Nevertheless, the simplest and most common AT approach in NLP is to augment the training set with textual adversarial examples $\mathbf{\tilde x}$ for each clean sample $\mathbf x$ by applying the attack to a model trained in the standard manner (Equation \ref{eqn:standard}).
% We refer to the above three groups of AT approaches as: \textit{Embedding-level}; \textit{Embedding Region-based AT}; and \textit{Text-based Augmentation AT} respectively.

\subsection{Model Calibration}\label{sec:back-cal}

Modern deep learning models are often miscalibrated, where the model's confidence in the predicted class does not reflect the ground truth correctness likelihood~\citep{DBLP:journals/corr/GuoPSW17}. A model with a predicted class confidence $P_{\hat{\theta}}(\hat c|\mathbf x)$, is defined as perfectly calibrated when
\begin{equation}
    P\left(\hat c = c^* | P_{\hat{\theta}}(\hat c |\mathbf x ) = p\right ) = p, \quad \forall p\in[0,1],
\end{equation}
where $\hat c =\mathcal F(\mathbf x; \hat{\theta})$ is the predicted class and the true (label) class is $c^*$. Any deviation from this indicates miscalibration. Typical single-value summaries for the calibration error are the Expected Calibration Error (ECE) and the Maximum Calibration Error (MCE)~\citep{Naeini2015ObtainingWC}.

% The extent of a model's miscalibration can be visualized on a reliability diagram~\citep{Degroot1983TheCA, pred-nicu}, displaying the sample accuracy as a function of model confidence. Any deviation from the identity line on a reliability diagram indicates miscalibration. Typical single-value summaries for the calibration error are the Expected Calibration Error (ECE) and the Maximum Calibration Error (MCE)~\citep{Naeini2015ObtainingWC}. 

% Although, the calibration error is often measured using the Expected Calibration Error or the Maximum Calibration Error, a popular and standard indicator of a model's extent of miscalibration is given by the Negative Log Likelihood (NLL)~\citep{Hastie-Friedman-Tisbshirani-2017},
% \begin{equation}\label{eqn:nll}
%     \mathcal L_{\texttt{NLL}} = -\sum_i\log  P_{\hat{\theta}}(c_i^*|\mathbf x_i )
% \end{equation}
% measured for a held-out validation set $\{\mathbf x_i, c^*_i\}$. In deep learning literature, the NLL is referred to as the cross entropy loss.

\subsection{Obfuscated Gradients}
Computer vision literature has demonstrated that a `false sense of security' to adversarial attacks on image classification systems can arise due to `obfuscated gradients’~\citep{DBLP:journals/corr/abs-1802-00420, 10.1145/3052973.3053009, tramer2018ensemble}. Obfuscated gradients block an adversary’s search process for an adversarial example and hence give the sense that the system is robust to the adversarial attack --- however, this is not the case, as adversarial examples still exist, but specific gradient-based mechanisms used by adversaries to find the adversarial examples are not effective when obfuscated gradients are present. In computer vision, obfuscated gradients in typical image systems can arise due to various reasons, including `shattered gradients’, `stochastic gradients’, exploding/vanishing gradients and any form of gradient masking~\citep{DBLP:journals/corr/abs-1802-00420, 10.1145/3052973.3053009, tramer2018ensemble}. However, to our knowledge, no previous work has explored how the illusion of robustness phenomenon emerges in NLP systems.

In this work, we observe that in various adversarial training regimes designed for NLP tasks, there is often the risk that systems become extremely miscalibrated. Extreme miscalibration effectively results in `obfuscated gradients’, which in turn results in the Illusion of Robustness. Hence, in summary, both computer vision and NLP systems can suffer from the Illusion of Robustness due to obfuscated gradients, but the practical scenarios which cause these obfuscated gradients differ for image and NLP classification systems. As a result, the methods used to mitigate the root causes behind obfuscated gradients for image classification and NLP classification systems, also differ.
\section{The Illusion of Robustness} \label{sec:conf}

Certain modeling approaches can lead to an illusion of robustness (IOR). In computer vision, it is shown that obfuscating gradients \citep{DBLP:journals/corr/abs-1802-00420} through \textit{shattered}, \textit{stochastic} or \textit{exploding/vanishing} gradients; or masking gradients \citep{10.1145/3052973.3053009, tramer2018ensemble} can lead to such an IOR - the model appears robust to adversarial attacks. We build on these works and argue that (un)intentional extreme model miscalibration is an example of a realistic cause of IOR.

The robustness gains observed for traditional NLP AT approaches (Equation \ref{eqn:robust}), may not always be due to inherent robustness gains, but can be a consequence of extreme model miscalibration. This miscalibration can induce extreme confidence predictions, such that the model's predicted class confidence $P_{\hat{\theta}}(\hat c|\mathbf x)$ is either very high (overconfident) or very low (underconfident). Figure \ref{fig:extreme} (using a standard NLP model, test dataset and adversarial attack described in Section \ref{sec:exps}) demonstrates that highly miscalibrated models with extreme confidence values in the predicted class (around 1.0 for overconfident models or $1/C$, where $C$ is the number of classes for underconfident models) are significantly more ``robust'' to adversarial attacks.

This apparent increase in robustness of extremely miscalibrated models can be explained. For both underconfident and overconfident models, the predicted class confidence has very little variance for different input sequences, $\mathbf x$,
\begin{equation}
   E_{p(\mathbf x)} [P_{\hat{\theta}}(\hat c|\mathbf x) - \mathbb E_{p(\mathbf x)} [P_{\hat{\theta}}(\hat c|\mathbf x)]]^2 < \zeta,
\end{equation}
where $\zeta$ is some small variance. The narrow confidence distribution makes it challenging for an adversary to identify an appropriate search direction for adversarial examples. To illustrate this, consider a miscalibrated model with extremely high confidence in the predicted class probability, $P_{\hat{\theta}}(\hat c|\mathbf x) \approx 1.0$, then for most search directions $\mathbf d$ that are not in an adversarial direction  $\mathbf d \neq \tilde{\mathbf d}$ (where $\tilde{\mathbf x} = \mathbf x + \tilde{\mathbf d}$) the model has very little sensitivity,\footnote{Note that these strict mathematical operations are not defined for the input text space and are simply representative of equivalent discrete textual space perturbations.} i.e.,
\begin{equation} \label{eqn:block}
    \mathbf d^T\nabla_{\mathbf x}P_{\hat{\theta}}(\hat c|\mathbf x)\approx 0.
\end{equation}
Consequently, any whitebox adversarial attack approach looking to exploit gradients or even a blackbox attack approach measuring the sensitivity of the predicted probability, has a small confidence range to observe. The implication is  that the impact of any proposed perturbation gives a very \textit{noisy} signal to its actual effect on the output. As a result, the adversarial attack search process will converge extremely slowly or outright fail to find the desired adversarial perturbation direction $\tilde{\mathbf d}$. We verify this hypothesis empirically in Appendix \ref{sec:noisy-grad}, corroborating related computer vision literature on gradient obfuscation~\citep{DBLP:journals/corr/abs-1802-00420}. We first demonstrate how to induce an IOR by explicitly miscalibrating models at test-time, before discussing how this can happen unintentionally during training.

%To demonstrate that miscalibration can cause an apparent increase in robustness against of-the-shelf adversarial attacks, we consider models that are explicitly induced with overconfidence or underconfidence (Section \ref{sec:explicit}). We further demonstrate that standard AT approaches can also implicitly induce extreme confidence and also cause an apparent increase in robustness (Section \ref{sec:implicit}). Section \ref{sec:mitigate} shows that this increase in robustness is an illusion of robustness (IOR).

\subsection{Explicit: Test-time Temperature Scaling} \label{sec:explicit}

Let $\hat{\theta}$ be a model trained using the standard training objective, as in Equation \ref{eqn:standard}. For this model with predicted logits, $l_1, \hdots, l_C$ for $C$ output classes, the probability of a specific class is typically estimated by the Softmax function, $P_{\hat{\theta}}(c|\mathbf x) = \frac{\exp{(l_c)}}{\sum_i\exp{(l_i)}}$. However, we can intentionally miscalibrate the model and increase the model confidence at \textit{test time} by using a temperature, $T=T_d$, to scale the predicted logits,
\begin{equation}\label{eqn:temp}
    P_{\hat{\theta}}(c|\mathbf x; T) = \frac{\exp{(l_c/T)}}{\sum_i\exp{(l_i/T)}}.
\end{equation}
A design choice of $T_d\ll 1.0$ concentrates the probability mass in the largest logit class to create an \textit{overconfident} model, whilst conversely $T_d\gg1.0$ creates an \textit{underconfident} model. Hence, explicitly setting a design temperature $T^{(d)}$ at inference time can be used to serve highly miscalibrated models, which can disrupt an adversary's attack search process (Equation \ref{eqn:block}), whilst maintaining the simplicity of the standard training objective (Equation \ref{eqn:standard}).

\subsection{Implicit Overconfidence: Grad.~Norm.}\label{sec:implicit}

%Section \ref{sec:explicit} presents an explicit temperature scaling method to generate a highly miscalibrated system, which cause an illusion of robustness for out-of-the-box adversarial attacks. However, it is possible that
Having shown how to induce an IOR by explicitly scaling a model's temperature, we now discuss how certain implementation strategies and algorithmic features in adversarial training (AT) procedures can also implicitly induce model overconfidence.

We first examine the recently proposed Danskin Descent Direction approach for adversarial training (DDi-AT;~\citealp{latorre2023finding}). \citet{latorre2023finding} adapted the standard AT paradigm (Equation \ref{eqn:robust}) to identify optimal gradient update directions for increased robustness, showing promising results in computer vision. We describe the NLP-specific implementation in Appendix \ref{sec:ddi}. 

In preliminary experiments, we observed that DDi-AT creates highly overconfident models without compromising clean accuracy, such that a DDi-AT model almost always predicts with near 100\% confidence in its predicted class, $P_{\hat{\theta}}(c|\mathbf x)\approx 1.0$ (Table \ref{tab:conf-comp}). Further ablations, detailed in Appendix \ref{sec:gradnorm}, revealed that the gradient normalization step in the DDi algorithm was responsible for this model overconfidence. In the following section, we verify this by applying gradient normalization to other AT schemes that may use it during training.

\subsection{Experiments} \label{sec:exps}
We first present the experimental setup for all experiments in this paper before showing how illusions of robustness can arise when evaluating highly over-confident or under-confident models for robustness. 
%demonstrate how explicit or implicit training approaches that cause a model to become highly underconfident or overconfident (miscalibrated) suffer from an \textit{illusion of robustness} (IOR), where the models appear robust to adversarial attacks.
% We first demonstrate how explicit or implicit training approaches that cause a model to become highly underconfident or overconfident (miscalibrated) suffer from an \textit{illusion of robustness} (IOR), where the models appear robust to adversarial attacks. We then show how simple approaches can be used to pierce this illusion. Finally, we present a simple training modification that can give a genuine boost in adversarial robustness for standard training and popular NLP AT approaches.

%\subsubsection{Experimental Setup} \label{sec;exp-setup}

\paragraph{Data.} Experiments are carried out on six standard NLP classification datasets. For IOR experiments we use Rotten Tomatoes~\citep{Pang+Lee:05a}; the Twitter Emotions Dataset~\citep{saravia-etal-2018-carer}; and the AGNews dataset~\citep{Zhang2015CharacterlevelCN}. We observe the same general trends across all datasets, and therefore present the results on Rotten Tomatoes here and include the others in Appendix \ref{sec:other-data}. %\footnote{Equivalent results for the other datasets are presented in Appendix \ref{sec:other-data} for IOR experiments and in Appendix \ref{sec:ht-results} for high temperature training experiments. The same general trends are observed across the different datasets.} 

\paragraph{Models.} We follow existing adversarial robustness literature and use Transformer~\citep{DBLP:journals/corr/VaswaniSPUJGKP17} encoders, which are state-of-the-art on many classification tasks.~\footnote{Appendix \ref{sec:llms} demonstrates the superior performance of encoder-only models relative to generative language models for many classification tasks.} Specifically, we consider the base variants of DeBERTa~\citep{DBLP:journals/corr/abs-2006-03654}, RoBERTa~\citep{liu2019roberta} and BERT \citep{DBLP:journals/corr/abs-1810-04805}. We observe the same general trends across all models, and therefore present the results for DeBERTa here and the others in Appendix \ref{sec:other-models}.%\textbf{The results in the main paper are presented for the DeBERTa model}.\footnote{Equivalent results presented for the other models in Appendix \ref{sec:other-models} for IOR and Appendix \ref{sec:models} for high temperature training. Identical trends are observed for all the models. Hyperparameter settings for training of these models are given in Appendix \ref{sec:hyperparam}. All experiments are run over three random seeds.}

\begin{table}[t]
    \centering
    \small
    \fontsize{8}{11}\selectfont
    \begin{tabular}{l|l|c|cc}
    \toprule
    \textbf{Intv.~Type} &
        \textbf{Method} & \textbf{clean} &  \textbf{$\bar P(\hat c|\mathbf{x}_{\text{clean}})$} & \textbf{$\bar P(\hat c|\mathbf{x}_{\text{adv}})$} \\ \midrule
        \multirow{3}{*}{None} & baseline & $\underset{\pm 0.30}{88.96}$ & $\underset{\pm0.26}{97.08}$ & $\underset{\pm0.68}{86.04}$\\
        &pgd & $\underset{\pm0.73}{88.24}$ & $\underset{\pm0.42}{97.56}$ & $\underset{\pm0.59}{87.25}$\\
        & ascc & $\underset{\pm0.36}{87.77}$ & $\underset{\pm0.24}{97.50}$&$\underset{\pm0.43}{86.99}$\\        
        \midrule
        \multirow{2}{*}{Explicit} & $\downarrow$conf & $\underset{\pm0.30}{88.96}$ & $\underset{\pm0.00}{50.00007}$ & $\underset{\pm0.00}{50.00004}$\\
        & $\uparrow$conf & $\underset{\pm0.30}{88.96}$ & $\underset{\pm0.02}{99.98}$ & $\underset{\pm0.01}{99.95}$\\ \midrule
        \multirow{4}{*}{Implicit} & baseline$^*$ & $\underset{\pm0.19}{88.56}$ & $\underset{\pm0.04}{99.96}$ & $\underset{\pm0.02}{99.93}$\\
        & ddi-at & $\underset{\pm0.49}{87.90}$ & $\underset{\pm0.03}{99.97}$ & $\underset{\pm0.01}{99.91}$\\
        & pgd$^*$ & $\underset{\pm0.64}{88.59}$ & $\underset{\pm0.04}{99.96}$ & $\underset{\pm0.01}{99.90}$\\
        & ascc$^*$ & $\underset{\pm0.36}{87.77}$ & $\underset{\pm0.04}{99.97}$ & $\underset{\pm0.01}{99.92}$\\
        \bottomrule
    \end{tabular}
    \caption{Clean accuracy (\%) and model confidence (\%) on clean and adv.~examples from Rotten Tomatoes for extreme confidence systems. We categorize them by intervention type --- explicit temperature scaling and implicit induction via gradient normalization during training. We use pwws to generate the adv.~examples.}
    \label{tab:conf-comp}
\end{table}

\paragraph{Adversarial attacks.} We experiment with four common adversarial attacks. BERT-based Adversarial Examples (\textit{bae})~\citep{garg-ramakrishnan-2020-bae} is a word-level blackbox attack, where the adversary only has access to the model inputs and predictions. We also include Textfooler (\textit{tf})~\citep{DBLP:journals/corr/abs-1907-11932} and Probability Weighted Word Saliency (\textit{pwws})~\citep{ren-etal-2019-generating}, more powerful word-level attacks. Finally, we include the DeepWordBug (\textit{dg})~\citep{DBLP:journals/corr/abs-1801-04354} attack as a whitebox, \textit{character}-level adversarial attack. We use the TextAttack implementations and their default settings ~\citep{morris2020textattack}. To measure the impact of each adversarial attack, we report the \textit{adversarial accuracy}, which is the accuracy of the target model on adversarial examples found by the attack.

\paragraph{Explicit temperature scaling.} We create under- and over-confident models, $\downarrow$\textit{conf} and $\uparrow$\textit{conf}, by scaling the temperature as in Section \ref{sec:explicit}.

\paragraph{AT approaches.} We consider a range of AT methods: Danskin Descent Direction (\textit{ddi-at};~\citealp{latorre2023finding}), PGD-K~(\textit{pgd}; \citealp{madry2018towards}) and FreeLB~(\textit{freelb}; \citealp{Zhu2020FreeLB} as embedding-space AT schemes, and ASCC~(\textit{ascc}; \citealp{DBLP:journals/corr/abs-2107-13541}) as a text-embedding combined AT approach. We further create variants of the baseline, \textit{pgd} and \textit{ascc} that use gradient normalization during training, named \textit{baseline}$^*$, \textit{pgd}$^*$, and \textit{ascc}$^*$, respectively. Finally, we consider the most common NLP AT method: augmenting the training set with adversarial examples, using DeepWordBug as the representative attack (\textit{dg-aug}). Hyperparameters are detailed in Appendix \ref{sec:hyperparam}.

\subsubsection{Results}

\paragraph{Model Confidence.} We see from Table \ref{tab:conf-comp} that the $\uparrow$\textit{conf}, \textit{baseline}$^*$, \textit{ddi-at}, \textit{pgd}$^*$, and \textit{ascc}$^*$ models are significantly more confident than the models with no intervention, whilst the $\downarrow$\textit{conf} model is far less confident, as intended. Note that the differences in the confidence are more pronounced for the adversarial examples (\textit{pwws} is used to attack the test set). 
%Despite these differences, the clean accuracy on the test data is the same or similar to that of the baseline model. 
The differences in confidence between the original and gradient normalization variants of the \textit{baseline}, \textit{pgd}, and \textit{ascc} models verifies our hypothesis that using gradient normalization during training is one cause of extreme overconfidence.

\paragraph{Robustness.}
Table \ref{tab:attack} presents the adversarial robustness of each model as measured by the adversarial accuracy under the different adversarial attacks. 
%For comparison, we include the AT approaches (dg-aug, pgd, ascc, freelb), which have been designed to \textbf{not} be overconfident by removing gradient normalization during training (Appendix \ref{sec:cal-base}). In general, the baseline AT approaches (dg-aug, pgd, ascc, freelb) do increase model robustness across all the different attack methods, with the augmentation approach being the most effective. The low confidence model also demonstrates comparable adversarial robustness to the augmentation-based approach. 
While the regular AT approaches (\textit{dg-aug}, \textit{pgd}, \textit{ascc}, \textit{freelb}) increase robustness to some extent, the gains in robustness of the highly overconfident models ($\uparrow$\textit{conf}, \textit{ddi-at}, \textit{baseline}$^*$, \textit{pgd}$^*$, \textit{ascc}$^*$) \textbf{appear} to outstrip them by two- to three-fold.
% The increased model robustness is particularly surprising for the explicit confidence manipulation models, $\downarrow$conf and $\uparrow$conf, as the models are identical to \textit{base} (which has not been trained on any adversarial examples), with the only change being temperature scaling of the logits. 

\begin{table}[t]
    \centering
    \small
    \fontsize{8}{11}\selectfont
    \begin{tabular}{l|c|cccc}
    \toprule
      \textbf{ Method}  & \textbf{clean} & \textbf{bae} & \textbf{tf} & \textbf{pwws} & \textbf{dg} \\ \midrule
       
        baseline  &  $\underset{\pm0.30}{88.96}$ & $\underset{\pm1.20}{31.39}$ & $\underset{\pm0.49}{17.82}$ & $\underset{\pm0.62}{20.42}$ & $\underset{\pm0.94}{20.11}$\\ 
        \midrule
        dg-aug & $\underset{\pm0.39}{87.12}$ &$\underset{\pm1.59}{34.74}$&$\underset{\pm1.83}{22.36}$&$\underset{\pm2.57}{26.11}$&$\underset{\pm0.75}{37.43}$  \\
        
        pgd & $\underset{\pm0.73}{88.24}$ &$\underset{\pm0.57}{33.65}$&$\underset{\pm0.47}{19.92}$&$\underset{\pm0.87}{26.70}$&$\underset{\pm0.61}{26.05}$\\
        
        ascc & $\underset{\pm0.36}{87.77}$ &$\underset{\pm0.64}{33.61}$&$\underset{\pm2.17}{15.13}$&$\underset{\pm0.77}{23.50}$&$\underset{\pm2.11}{26.80}$\\
        
        freelb & $\underset{\pm0.32}{88.74}$ &$\underset{\pm0.52}{32.52}$&$\underset{\pm1.70}{19.51}$&$\underset{\pm0.70}{24.55}$&$\underset{\pm0.73}{24.52}$\\\midrule
        
        $\downarrow$conf (\S \ref{sec:explicit}) & $\underset{\pm0.30}{88.96}$ & $\underset{\pm0.94}{31.21}$ & $\underset{\pm0.99}{20.98}$ & $\underset{\pm0.89}{25.17}$ & $\underset{\pm2.78}{32.18}$\\

         $\uparrow$conf (\S \ref{sec:explicit}) & $\underset{\pm0.30}{88.96}$ & $\underset{\pm1.18}{37.71}$ & $\underset{\pm0.73}{54.35}$ & $\underset{\pm0.62}{59.29}$ & $\underset{\pm1.81}{65.60}$\\
        \midrule

        baseline$^*$ & $\underset{\pm0.19}{88.56}$ &   $\underset{\pm3.57}{33.71}$ &   $\underset{\pm5.01}{47.22}$ &   $\underset{\pm3.03}{53.03}$ &   $\underset{\pm4.09}{59.10}$\\
        
        ddi-at (\S\ref{sec:implicit}) & $\underset{\pm0.49}{87.90}$ &  $\underset{\pm0.75}{39.18}$ &  $\underset{\pm1.67}{56.54}$ &  $\underset{\pm0.99}{61.07}$ &  $\underset{\pm1.01}{66.73}$\\ 

        pgd$^*$ & $\underset{\pm0.64}{88.59}$& $\underset{\pm0.55}{39.94}$ & $\underset{\pm1.04}{58.02}$  & $\underset{\pm0.77}{64.45}$ & $\underset{\pm0.83}{67.02}$\\
        
        ascc$^*$ & $\underset{\pm0.36}{87.77}$ & $\underset{\pm0.69}{40.01}$ & $\underset{\pm1.57}{54.32}$ & $\underset{\pm0.86}{63.99}$ & $\underset{\pm0.93}{67.43}$ \\   
        \bottomrule
    \end{tabular}
    \caption{Accuracy (\%) of extreme confidence systems on Rotten Tomatoes compared to standard AT methods under various adversarial attacks.}
    \vspace{-0.1em}
    \label{tab:attack}
\end{table}

We argue that the apparent increase in robustness of the extreme confidence models ($\downarrow$\textit{conf}, $\uparrow$\textit{conf}, \textit{baseline}$^*$, \textit{ddi-at}, \textit{pgd}$^*$ and \textit{ascc}$^*$) in Table \ref{tab:attack} is due to the attack search process being disrupted, but the models are still susceptible to adversarial examples. We know this must be true for the explicitly temperature-scaled models since the predicted class never changes from the \textit{baseline}'s, only its confidence. We further empirically find that extreme confidence results in a noisier search for the adversarial attack. We discuss this in greater detail in Appendix \ref{sec:noisy-grad} due to space limitations.

%\footnote{Appendix \ref{sec:noisy-grad} empirically shows that extreme confidence results in a noisier search for regular adversarial attacks.} i.e. the models are actually susceptible to adversarial examples but the adversarial attacks are unable to find these adversarial examples. Hence, the observed robustness is an IOR. 
\section{Piercing the Illusion} \label{sec:mitigate}

%Section \ref{sec:conf} demonstrates how intentional or accidental extreme miscalibration of a model can disrupt out-of-the-box adversarial attack search methods by obfuscating gradients and thus giving an apparent gain in robustness. This section highlights that the observed gain in robustness is an illusion of robustness (IOR), as we propose simple approaches that an adversary can use to mitigate extreme model confidences to remove the disruption to the attack search methods.

Having demonstrated how to induce an IOR from the developer or defender's perspective, we now discuss how to nullify it as an adversary. Although the following approaches involve modifying aspects of the model's output at test-time, these modifications are only used to create/find adversarial examples, which can then be applied to the original, un-modified model served by the model developer.

\subsection{Naive Test-Time Temperature Calibration} \label{sec:cal}

Highly miscalibrated models (Section \ref{sec:conf}) interfere with an adversarial attack's ability to find meaningful search directions due to the little sensitivity in the predicted probabilities. An adversary aims to mitigate this disruption to the attack search process. The natural solution, then, is to calibrate the model so that the confidences are in a sensible range and can now be exploited by adversarial attacks.

A strong indicator of model miscalibration (Section \ref{sec:back-cal}) can be given by the Negative Log Likelihood (NLL;~\citealp{Hastie-Friedman-Tisbshirani-2017}). Thus, assuming access to the output model logits $l_1, \hdots, l_C$ and a labelled validation set of data $\{\mathbf x_i, c^*_i\}_i$, an adversary can apply test-time temperature calibration~\citep{DBLP:journals/corr/GuoPSW17}. This works regardless of \textit{how} the model was miscalibrated (implicitly or explicitly). Now, the adversary optimizes an adversarial temperature, $T_a$ to minimize the Negative Log Likelihood (NLL) of the validation set samples, 
\begin{equation}\label{eqn:nll-ta}
    T_a = \argmin_{T} \sum_i -\log P_{\hat{\theta}}(c^*_i|\mathbf x_i; T),
\end{equation}
where $P_{\hat{\theta}}(c^*|\mathbf x; T)$ is the confidence of the true class after temperature scaling as in Equation \ref{eqn:temp}. Due to the continuous nature of the transformation and the need to optimize a single parameter, $T_a$, we use standard gradient descent optimization.\footnote{Inspired by \url{https://github.com/gpleiss/temperature_scaling/tree/master}.}

Other than temperature optimization, an adversary can attempt other post-training model calibration approaches such as Histogram Binning~\citep{10.5555/645530.655658}, isotonic regression~\citep{10.1145/775047.775151} and multi-class versions of Platt scaling~\citep{pred-nicu, Platt2007ProbabilisticOF}. However, we empirically find temperature calibration to be the most practical and effective for an adversary seeking to mitigate a model's IOR. We discuss this in greater detail in Appendix \ref{sec:cal-compare}.

\subsection{Adversarial Temperature Optimization}\label{sec:opt}

While simple, the naive calibration approach has two shortcomings:
\begin{enumerate}
    \item The adversarial temperature, $T_a$ is not directly tuned to minimize adversarial robustness, as it only considers the likelihood of clean examples in a validation set.
    \item Learning the adversarial temperature, $T_a$ to minimize the NLL (Equation \ref{eqn:nll-ta}) uses a gradient descent--based optimization algorithm that is sensitive to hyperparameters and does not guarantee an optimal solution.
\end{enumerate}

Hence, we now outline an algorithm that directly optimizes the adversarial temperature $T_a$ to minimize a model's adversarial robustness at test time. We define the adversarial accuracy, $\mathcal Q()$ as a function of the temperature parameter,
\begin{equation} \label{eqn:adv-acc}
    \mathcal Q(T)= \frac{1}{J}\sum_j\mathbb I \left[\mathcal F (\tilde{\mathbf x}_j(T)) = c_j^*\right ],
\end{equation}
where $\tilde{\mathbf x}_j(T)$ represents the adversarial example generated from an adversarial attack on the given model, $\hat\theta$ with the logits scaled by a temperature $T$ as in Equation \ref{eqn:temp}. Figure \ref{fig:extreme} illustrates that as the temperature parameter is swept from large to small values (increasing model confidence), the adversarial accuracy, $\mathcal Q()$ behaves almost as a convex function of temperature, $T$, such that, $\mathcal Q(\alpha T_1 + (1-\alpha)T_2)\leq \alpha\mathcal Q(T_1) + (1-\alpha)\mathcal Q(T_2)$, where $0\leq\alpha\leq 1$. The optimal adversarial temperature $T_a$ minimizes the adversarial accuracy $\mathcal Q(T)$,
\begin{equation} \label{eqn:adv-opt}
    T_a = \argmin_{T}\mathcal Q(T).
\end{equation}
Hence, $T_a$ can be found efficiently over the non-differentiable convex function, $\mathcal Q()$, using search methods such as golden section search~\citep{Kiefer-1953}. We use the Brent-Dekker method, an extension of golden section search that accounts for a parabolic convergence point~\citep{10.1093/comjnl/14.4.422}. 
% Note, as is the case for the calibration approach of Section \ref{sec:cal}, to optimize for $T_a$, an adversary is not required to query the target model multiple times as the adversary only requires the output model logits $l_1, \hdots, l_C$.

%one may adopt the naive temperature calibration approach to pierce the IOR when computational resources are scarce, but use the optimization approach outlined here otherwise.
% The greatest computational cost can be attributed to calculation of the adversarial accuracy (Equation \ref{eqn:adv-acc}), as this requires an adversarial attack to be applied to each clean sample in the validation set, $\{\mathbf x_j, c^*_j\}_{j=1}^J$.

\subsection{Experiments}
\paragraph{Setup.} We maintain the experimental setup as Section \ref{sec:exps} and supplementary results for different models and datasets are provided in Appendix \ref{sec:app-ior}.

\paragraph{IOR mitigation.}
We experiment with the two proposed approaches to mitigate the disruption of the adversarial attack search processes and remove the IOR: naive temperature calibration (\textit{cal}) and adversarial temperature optimization (\textit{opt}). The learning rate is set to 0.01 with a maximum of 5000 iterations for \textit{cal}. For \textit{opt}, we use DeepWordBug to attack the validation set when optimizing for $T_a$. 

\paragraph{Evaluation.} We first modify the model by scaling the predicted logits by $T_a$ and then the adversarial attacks are run on the modified model to find adversarial examples. The original, unmodified model is lastly evaluated on these adversarial examples.

\begin{table}[t]
    \centering
    \small
    \fontsize{7.5}{11}\selectfont
    \begin{tabular}{lc|c|cccc}
    \toprule
       \textbf{Method} & \textbf{Adv.} & \textbf{clean} & \textbf{bae} & \textbf{tf} & \textbf{pwws} & \textbf{dg} \\ \midrule
        baseline  & -& $\underset{\pm0.30}{88.96}$ & $\underset{\pm1.20}{31.39}$ & $\underset{\pm0.49}{17.82}$ & $\underset{\pm0.62}{20.42}$ & $\underset{\pm0.94}{20.11}$\\ \midrule

        $\downarrow$conf &-& $\underset{\pm0.30}{88.96}$ & $\underset{\pm0.94}{31.21}$ & $\underset{\pm0.99}{20.98}$ & $\underset{\pm0.89}{25.17}$ & $\underset{\pm2.78}{32.18}$\\
        &cal& $\underset{\pm0.30}{88.96}$ & $\underset{\pm0.34}{31.52}$ & $\underset{\pm0.43}{21.89}$ & $\underset{\pm1.31}{27.58}$ & $\underset{\pm0.34}{31.52}$\\
        &opt & $\underset{\pm0.30}{88.96}$ & $\underset{\pm1.15}{31.44}$ & $\underset{\pm0.49}{17.82}$ & $\underset{\pm0.64}{20.86}$ & $\underset{\pm1.66}{21.98}$\\ \midrule 
        
        $\uparrow$conf &-& $\underset{\pm0.30}{88.96}$ & $\underset{\pm1.18}{37.71}$ & $\underset{\pm0.73}{54.35}$ & $\underset{\pm0.62}{59.29}$ & $\underset{\pm1.81}{65.60}$\\
        &cal & $\underset{\pm0.30}{88.96}$ & $\underset{\pm1.20}{31.39}$ & $\underset{\pm0.49}{17.82}$ & $\underset{\pm0.74}{20.45}$ & $\underset{\pm1.46}{21.64}$\\
        &opt & $\underset{\pm0.30}{88.96}$& $\underset{\pm1.20}{31.39}$& $\underset{\pm0.49}{17.82}$& $\underset{\pm0.94}{20.90}$& $\underset{\pm0.82}{21.06}$\\ \midrule

            baseline$^*$ && $\underset{\pm0.19}{88.56}$ &   $\underset{\pm3.57}{33.71}$ &   $\underset{\pm5.01}{47.22}$ &   $\underset{\pm3.03}{53.03}$ &   $\underset{\pm4.09}{59.10}$\\

            & cal & $\underset{\pm0.19}{88.56}$ & $\underset{\pm0.14}{32.40}$ & $\underset{\pm0.48}{18.79}$ & $\underset{\pm1.22}{21.36}$ & $\underset{\pm0.66}{21.11}$ \\ \midrule

        ddi-at &-& $\underset{\pm0.49}{87.90}$ &  $\underset{\pm0.75}{39.18}$ &  $\underset{\pm1.67}{56.54}$ &  $\underset{\pm0.99}{61.07}$ &  $\underset{\pm1.01}{66.73}$\\
        &cal& $\underset{\pm0.49}{87.90}$ &  $\underset{\pm0.57}{31.80}$ &  $\underset{\pm3.01}{18.36}$ &  $\underset{\pm1.96}{23.08}$ &  $\underset{\pm3.38}{22.89}$\\
        &opt &  $\underset{\pm0.49}{87.90}$ & $\underset{\pm0.57}{31.80}$ & $\underset{\pm3.32}{18.88}$ & $\underset{\pm1.03}{22.16}$ & $\underset{\pm1.12}{22.28}$\\ \midrule

        pgd$^*$ & - & $\underset{\pm0.64}{88.59}$& $\underset{\pm0.55}{39.94}$ & $\underset{\pm1.04}{58.02}$  & $\underset{\pm0.77}{64.45}$ & $\underset{\pm0.83}{67.02}$ \\
        & cal & $\underset{\pm0.64}{\underline{88.59}}$ &$\underset{\pm0.20}{33.71}$&$\underset{\pm0.86}{17.73}$&$\underset{\pm1.80}{25.20}$&$\underset{\pm1.46}{25.74}$\\ \midrule

        ascc$^*$ & - & $\underset{\pm0.36}{87.77}$ & $\underset{\pm0.69}{40.01}$ & $\underset{\pm1.57}{54.32}$ & $\underset{\pm0.86}{63.99}$ & $\underset{\pm0.93}{67.43}$ \\
        & cal & $\underset{\pm0.36}{87.77}$ &$\underset{\pm0.64}{33.61}$&$\underset{\pm2.17}{15.13}$&$\underset{\pm0.77}{23.50}$&$\underset{\pm2.11}{26.80}$\\
        
        \bottomrule
    \end{tabular}
    \caption{Clean and adv.~accuracy (\%) on Rotten Tomatoes after mitigating the \textit{illusion of robustness} of highly miscalibrated systems using temperature calibration (\textit{cal}) or adversarial temperature optimization (\textit{opt}).}
    \label{tab:cal}
\end{table}

\subsubsection{Results}
Table \ref{tab:cal} shows the impact of the different adversarial approaches (\textit{cal} and \textit{opt}) to learn $T_a$ on the adversarial robustness of the models. For the overconfident models, $\uparrow$\textit{conf}, \textit{baseline}$^*$, \textit{ddi-at}, \textit{pgd}$^*$ and \textit{ascc}$^*$, simple temperature calibration (\textit{cal}) is sufficient to cause a significant drop in model robustness.\footnote{Appendix \ref{sec:cal-error} discusses the relationship between the calibration error and the model confidence.} For the $\downarrow$\textit{conf} model, the temperature optimization approach (\textit{opt}) is necessary to significantly reduce robustness, illustrating its efficacy at nullifying IORs.

It is worth analyzing why simple calibration (\textit{cal}) is effective in removing IOR in the overconfident models ($\uparrow$conf) but not for the underconfident models ($\downarrow$conf). This observation can perhaps be explained by considering the search space for the calibrating temperature: the \textit{cal} method is naive temperature calibration, where a temperature parameter $T_a$ (divisor of logits) is learnt on a validation set by minimizing the negative log likelihood. The solution for $T_a$ differs when calibrating low-confidence and high-confidence models. For low-confidence models, $T_a$ must lie between 0 and 1 (as the logits have to be scaled up), and for high-confidence models, $T_a$ only needs to be greater than 1. Hence, the solution space for $T_a$ is more constrained when calibrating low-confidence models, meaning it is more challenging for gradient based search methods to find the solution space for $T_a$. Therefore, \textit{cal} struggles more when calibrating the low confidence models.

Critically, our results highlight the risk of some AT approaches, whether by design or as an implementation detail, giving the illusion of robustness even if they do not yield genuinely robust models.

\subsection{Discussion}
Although adversarial temperature optimization is more effective in nullifying the IOR, it is significantly slower than the naive calibration approach. Therefore, naive temperature calibration is favorable in computational resource--scarce settings and should always be run at minimum in robustness evaluations. Another consideration is detectability. When evaluating an API model, adversarial temperature optimization will require sending many similar queries as part of the attack, which has a higher chance of being detected and blocked by the API. In contrast, naive temperature calibration only requires querying the API for predictions on clean examples, which will appear far more innocuous.

One might even argue that to expose an IOR, it is unnecessary for an adversary to modify the model with adversarial temperature scaling to find adversarial examples. Instead, we could find adversarial examples for another model (e.g., \textit{baseline}) and transfer to the target model. We test this hypothesis in Appendix \ref{sec:ior-transfer}. We find that although the transfer attack from \textit{baseline} to \textit{ddi-at} reduces the adversarial accuracy, it is unable to bring it down to the same values as \textit{baseline}, as achieved by our proposed temperature scaling approaches.
\section{Raising the Training Temperature for Genuine Robustness} \label{sec:high-temp}

Section \ref{sec:conf} showed that it is easy to unintentionally develop AT schemes that do not yield true robustness gains but instead induce IOR. The success of temperature scaling at interfering with an attack's search process poses the natural question: Can a similar effect be induced in training such that it cannot be nullified, thus improving true robustness?

We now present a simple modification to standard training that boosts the true robustness of NLP models to unseen attacks. We consider an attack to be unseen when its adversarial examples are not used in adversarial training. In our experiments, this refers to all AT methods other than \textit{dg-aug}.

\subsection{Method}
 %By considering Deep Neural Networks (DNNs) in the computer vision domain, \citet{DBLP:journals/corr/PapernotMWJS15} demonstrate that knowledge distillation  \citep{hinton2015distilling, DBLP:journals/corr/abs-2104-07163} can be an effective training method. Although we suspect that part of the success of such a model distillation approach described by \citet{DBLP:journals/corr/PapernotMWJS15} is an IOR, there is an intuitive motivation behind aspects of distillation that could offer genuine robustness gains. In this work, we isolate the use of a high temperature during training in knowledge distillation as the main contributor to model robustness. In knowledge distillation a smaller model (student) is trained to replicate the behavior of a larger, more complex model (teacher), where a high temperature is often used to soften the probability distribution, giving a smoother decision boundary. Although the temperature parameter is not used to soften the teacher distribution in our context, 
 
Since scaling the temperature down at test time reduces an adversarial attack's efficacy, intuitively, we would like to ``bake'' this behavior into a model's weights such that it cannot be neutralized at the logit layer by the approaches from Section \ref{sec:mitigate}. 

We propose to do this by increasing the temperature during training, bringing the probabilities of the different classes closer together. This encourages the model's parameters to compensate by pushing the logits of the different classes further apart (Figure \ref{fig:logits} in the Appendix). This increases the distance to the class boundary in the logit space and makes it more difficult for an adversarial attack to change the predicted class~\citep{robey2023adversarial}. 

Adversarial robustness can also be viewed as type of generalization, and \citet{DBLP:journals/corr/abs-2010-07344} find that model generalization depends strongly on the training temperature, where larger temperatures yield stronger results for vision models. Therefore, our method can also be viewed as flattening the loss landscape, which has been shown to improve generalization for adversarial robustness in computer vision \citep{stutz2021relating}. Using a temperature greater than one during training has also seen success previously in neural machine translation to `smooth' the softmax distribution during training and prevent overfitting~\citep{dabre-fujita-2021-investigating}. Future work will aim to rigorously understand the observed empirical robustness gains of high temperature training.

%We therefore   The gain in robustness can be explained by considering the size of the class margin~\citep{robey2023adversarial}. A high temperature brings the probabilities of the different classes closer together, resulting in the model's parameters compensating by pushing the logits of the different classes further apart (demonstrated in Figure \ref{fig:logits} in the Appendix). This can be viewed as increasing the distance to the class boundary in the logit space and thus making it more difficult for an adversarial attack to change the predicted class. Future work will aim to rigorously understand and explain the observed empirical robustness gains of training with a high temperature.

 \subsection{Experiments}
 We maintain the setup from Section \ref{sec:exps}, with the addition of three datasets from the General Language Understanding Evaluation benchmark~(GLUE; \citealp{wang2019glue}): the Corpus of Linguistic Acceptability; Question-answering NLI; and the Microsoft Research Paraphrase Corpus. We further include gradient normalization ($^*$) in all experiments, as we found this to stave off decreases in clean accuracy as we increase training temperature (ablation is detailed in Appendix \ref{sec:heat-gradnorm-ablation}). To ensure the robustness gains are not IORs, we apply the test-time calibration described in Section \ref{sec:mitigate}. Appendix \ref{sec:heat-cal} demonstrates the extent of IOR when such test-time temperature scaling is not applied.

\begin{table}[t]
    \centering
    \small
    \begin{tabular}{l|c|cccc}
    \toprule
        Method & Clean & bae & tf & pwws & dg \\ \midrule
        
        baseline$^*$ &$\underset{\pm0.19}{\underline{88.56}}$ & $\underset{\pm0.14}{32.40}$ & $\underset{\pm0.48}{18.79}$ & $\underset{\pm1.22}{21.36}$ & $\underset{\pm0.66}{21.11}$  \\

         $\oplus\ T$ &$\underset{\pm0.44}{87.55}$ & $\underset{\pm0.84}{\underline{35.83}}$ & $\underset{\pm4.57}{\underline{26.83}}$ & $\underset{\pm3.07}{\underline{31.49}}$ & $\underset{\pm4.71}{\underline{35.18}}$  \\ \midrule

       % \textbf{ pwws-aug} &$\underset{\pm0.86}{87.24}$ &$\underset{\pm0.76}{35.26}$&$\underset{\pm0.95}{23.09}$&$\underset{\pm3.02}{35.70}$&$\underset{\pm0.88}{33.86}$  \\
       %   $\oplus$gT & $\underset{\pm0.34}{86.02}$ & $\underset{\pm0.38}{\textbf{37.27}}$ & $\underset{\pm0.44}{\textbf{31.49}}$ & $\underset{\pm0.81}{\textbf{40.84}}$ & $\underset{\pm0.62}{\textbf{39.81}}$ \\\midrule\midrule
        
        pgd$^*$&$\underset{\pm0.64}{\underline{88.59}}$ &$\underset{\pm0.20}{33.71}$&$\underset{\pm0.86}{17.73}$&$\underset{\pm1.80}{25.20}$&$\underset{\pm1.46}{25.74}$  \\
        $\oplus\ T$ & $\underset{\pm0.43}{87.77}$& $\underset{\pm0.33}{\underline{34.77}}$& $\underset{\pm1.76}{\underline{24.55}}$& $\underset{\pm1.08}{\underline{31.46}}$& $\underset{\pm2.64}{\underline{31.77}}$\\ \midrule
        
        freelb$^*$ &$\underset{\pm0.32}{\underline{88.74}}$ &$\underset{\pm0.52}{32.52}$&$\underset{\pm1.70}{19.51}$&$\underset{\pm0.70}{24.55}$&$\underset{\pm0.73}{24.52}$  \\

         $\oplus\ T$ & $\underset{\pm0.52}{88.02}$& $\underset{\pm0.80}{\underline{35.15}}$& $\underset{\pm0.96}{\underline{25.17}}$& $\underset{\pm0.68}{\underline{29.96}}$& $\underset{\pm1.04}{\underline{31.49}}$\\ \midrule

        ascc$^*$ &$\underset{\pm0.36}{\underline{87.77}}$ &$\underset{\pm0.64}{33.61}$&$\underset{\pm2.17}{15.13}$&$\underset{\pm0.77}{23.50}$&$\underset{\pm2.11}{26.80}$  \\

         $\oplus\ T$& $\underset{\pm0.80}{86.36}$& $\underset{\pm1.12}{\underline{34.93}}$& $\underset{\pm0.72}{\underline{27.36}}$& $\underset{\pm1.38}{\underline{30.93}}$& $\underset{\pm1.65}{\underline{33.46}}$\\ \midrule

        dg-aug$^*$ &$\underset{\pm0.39}{\underline{87.12}}$ &$\underset{\pm1.59}{34.74}$&$\underset{\pm1.83}{22.36}$&$\underset{\pm2.57}{26.11}$&$\underset{\pm0.75}{\underline{37.43}}$  \\

         $\oplus\ T$ & $\underset{\pm0.22}{87.09}$& $\underset{\pm2.64}{\underline{36.99}}$& $\underset{\pm2.86}{\underline{26.92}}$& $\underset{\pm1.67}{\underline{31.43}}$& $\underset{\pm1.90}{36.40}$\\ 

    \bottomrule
    \end{tabular}
    \caption{Adversarial Training (AT) combined with a training temperature of $T=200$ ($\oplus T$) on Rotten Tomatoes. We report the post-calibration (\textit{cal}) accuracy. The higher accuracy between the AT model and the AT$\oplus T$ model is underlined. The higher training temperature always improves robustness to unseen adversarial attacks.}
    \label{tab:AT-gradnorm-rt}
\end{table}

\begin{figure}[t]
    \centering
    \includegraphics[width=\columnwidth]{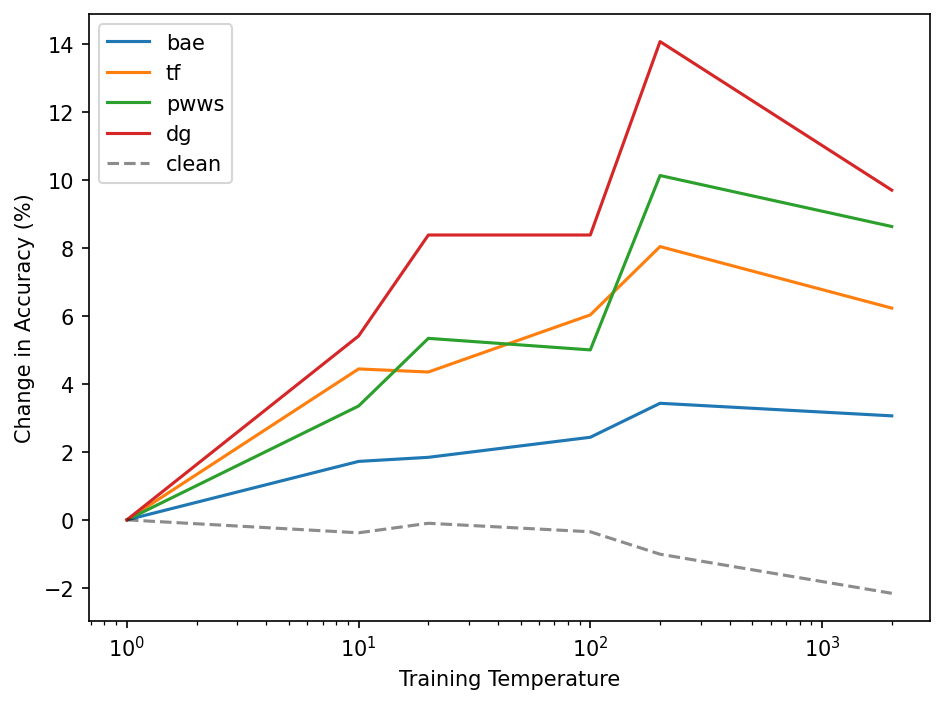}
    \caption{%The addition of a training temperature is a simple modification to standard model training, where the temperature parameter, $T$ is used to scale down the predicted model logits during training. 
    Change in post-calibration accuracy on Rotten Tomatoes as training temperature varies. We observe that a higher temperature during training increases robustness against unseen attacks (\textit{bae, tf, pwws, dg} here). The change in adversarial accuracy relative to the baseline ($T=1$) demonstrates the increase in robustness.}
    \label{fig:intro}
\end{figure}

\subsubsection{Results}
Results in Table \ref{tab:AT-gradnorm-rt} show that a high training temperature not only boosts genuine adversarial robustness for a model trained with the standard objective (\textit{baseline}$^*$$\oplus T$), but can also be combined with popular adversarial AT schemes for a further boost. We observe that the high training temperature approach improves the adversarial accuracy against unseen attacks for all of the adversarial training approaches experimented upon ($\oplus T$). This demonstrates that high temperature training is complementary to existing AT methods and consistently encourages a gain in genuine robustness. Interestingly, we observe that for \textit{dg-aug}$^*$, using a high temperature during adversarial training further improves adversarial accuracy for the unseen attacks but causes a slight drop in adversarial accuracy for the (seen) \textit{dg} attack compared to regular AT. This suggests that although a high training temperature successfully increases robustness to unseen attacks, it may not yield further robustness gains against seen attacks.
 
Additionally, Figure \ref{fig:intro} presents the change in clean and adversarial accuracy of a model trained with the standard training objective and different temperatures $T$ used during training.\footnote{Appendix \ref{app:heat-detailed-datasets} contains a detailed breakdown for each training temperature,  adversarial attack, and dataset.} We further observe a consistent robustness profile where robustness peaks at similar temperatures for all tested attack types. This is particularly useful for a model developer with access to only one type of adversarial attack. The training temperature can be tuned for optimal robustness on that specific attack form with the confidence that the robustness gains will transfer to the other unseen/unknown attack forms. Finally, we observe a decrease in accuracy at extremely large training temperatures, an indication that overly large temperatures may excessively smooth the predicted probability distribution and  make it too challenging for the model to learn. However, this is easy to avoid by sweeping the temperature with a single attack since there is a consistent robustness profile for each dataset.

\section{Conclusion}
NLP models are susceptible to adversarial attacks, where small changes in the input cause the model to predict incorrectly. Many adversarial training (AT) approaches have been proposed to induce robustness to adversarial attacks. In this work, we argued that the observed robustness gains may not be due to true increases in model robustness. We demonstrated how AT schemes can unknowingly create highly miscalibrated models that disrupt common adversarial attack search methods by obfuscating gradients, yielding up to three-fold perceived robustness gains. However, this is merely an \textit{illusion of robustness} (IOR). We proposed simple methods an adversary could use to circumvent such gains. Specifically, we showed how to perform test-time temperature scaling to mitigate disruptions to the adversarial attack search processes and pierce the IOR. Hence, we strongly recommend all adversarial robustness evaluations incorporate adversarial temperature scaling to ensure any observed robustness gain is genuine and not an \textit{illusion}. Finally, we proposed a practical training-time modification to increase a model's genuine robustness to unseen adversarial attacks and demonstrated its efficacy alone and in combination with other AT methods.

% \newpage

\section{Limitations}

\begin{itemize}
    \item Empirical results are presented for state-of-the-art encoder-based Transformer models. Although Appendix \ref{sec:llms} demonstrates that encoder-based models are more appropriate for many NLP classification tasks than the recently popularized generative Large Language Models (LLMs), it would be useful to investigate how susceptible these larger, decoder-based LLMs are to the IOR.
    \item We evaluate common Adversarial Training (AT) baselines to illustrate the IOR phenomenon. However, future work would benefit from conducting an in-depth study that includes other recently proposed approaches for adversarial robustness, e.g., contrastive learning based approaches~\citep{rim2021adversarial} and Textual Manifold Defence~\citep{nguyen-minh-luu-2022-textual}, where all inputs are mapped to a robust manifold. This would help the community understand the extent to which these proposed approaches are improving true robustness and the extent to which they may be unknowingly creating an IOR.
    \item In Section \ref{sec:high-temp}, we used a constant temperature during training for simplicity. However, varying the temperature over the course of training may further increase the effectiveness of high temperature training. Future work will aim to study the impact of temperature schedules.
\end{itemize}

\section{Risks and Ethics}

This work presents results on the topic of adversarial training. The contributions in this work encourage the development of truly robust systems and therefore there are no identified ethical concerns.

\bibliography{custom}
\bibliographystyle{acl_natbib}

\appendix

\newpage
\section{Danksin's Descent Direction for NLP} \label{sec:ddi}

\subsection{Original Theory}

\citet{latorre2023finding} demonstrate that the standard formulation and implementation of AT (as in Equation \ref{eqn:robust}) is potentially flawed. Specifically, solving the inner maximization to find the \textit{worst-case} adversarial example $\mathbf{\tilde x}$, can give a gradient direction (in standard stochastic gradient descent approaches), that can in fact \textit{increase} the robust loss (the new worst-case adversarial example, $\mathbf{\tilde x}$, with the updated model parameters, $\theta$, can give a robust loss that is greater than before the update step), i.e. worsening the adversarial robustness of the model. This flaw is attributed to the reliance on a single adversarial example, as a parameter gradient step to reduce the model's sensitivity to a particular adversarial example does not guarantee reduction in the model's sensitivity to all adversarial examples (the model may now be less robust to other adversarial examples) for a specific sample $\mathbf x$. The paper argues that their exist multiple solutions to the inner-maximization for the robust loss and the optimal parameter gradient direction depends on all of those solutions. Thus, Equation \ref{eqn:robust} can theoretically be adapted to selecting the adversarial example that maximises the gradient direction in each gradient update step for a batch size of $K$ samples,
\begin{align} \label{eqn:at-imp-ddi}
\theta_{i+1} = \Phi\left(\theta_i, \bm{\gamma}^* = -\frac{\nabla_{\theta}g(\mathbf x_{1:K}, \theta_i, \mathbf{\hat{\tilde x}}_{1:K})}{||\nabla_{\theta}g(\mathbf x_{1:K}, \theta_i, \mathbf{\hat{\tilde x}}_{1:K})||_2}\right), \nonumber \\ \nonumber g(\mathbf x_{1:K}, \theta_i, \mathbf{\hat{\tilde x}}_{1:K}) =\frac{1}{K}\sum_{k} \mathcal L(\mathbf{\hat{\tilde x}}_k, \theta_i),\\ \mathbf{\hat{\tilde x}}_k = \argmax_{\mathbf{\tilde x}\in\mathcal S^*(\theta_i, \mathbf x_k)}\left||\nabla_{\theta=\theta_i}\mathcal L(\mathbf{\tilde x}, \theta)\right||_2,
\end{align}
where $\Phi(\theta, \bm{\gamma})$ is the first-order stochastic gradient descent (SGD) algorithm used to update $\theta$ as per descent direction $\bm{\gamma}$, e.g. in standard SGD, $\Phi(\theta, \gamma) = \theta+\beta\bm{\gamma}$, where $\beta$ is the step-size (learning rate). Further $S^*(\theta_i, \mathbf x_k)$ represents the set of all maximizers of the robust loss,
\begin{equation}
    S^*(\theta, \mathbf x, \mathcal G) = \argmax_{\substack{\mathbf{\tilde x}:\\ \mathcal{G}(\mathbf x, \mathbf{\tilde x}, ) \leq\epsilon,\hspace{0.2em} \mathbf{\tilde x}\in\mathcal A}} \mathcal L(\mathbf{\tilde x}, \theta).
\end{equation}
This set of (robust loss) maximizers, $S^*(\theta, \mathbf x, \mathcal G)$ can theoretically be infinite. However, if assume we have access to a finite set with $M$ adversarial examples, such that they define,
\begin{equation} \label{eqn:s-m}
    S^{*(M)}(\theta, \mathbf x) = \{\mathbf{\tilde x}^{(1)}, \hdots, \mathbf{\tilde x}^{(M)}\},
\end{equation}
then \citet{latorre2023finding} propose an efficient algorithm termed, Danskin's Descent Direction (DDi), that provides a method to approximate the steepest direction, $\bm{\gamma}^*$ as though as if we are still selecting from the infinite set $S^*$~\footnote{Theorem 3 in the paper justifies the conditions to certify that the approximation is the steepest descent direction}, despite only having access to $S^{*(M)}$. The optimization problem over an infinite set in Equation \ref{eqn:at-imp-ddi} can be solved by finding an optimal linear combination, $\bm{\alpha}\in\triangle^M$ of the gradients of the loss, $\nabla_{\theta} g$ for each different adversarial example. Note that  $\triangle^M$ defines the $M$-dimensional simplex (on which $\bm{\alpha}$ lies). If we let $\nabla_{\theta}g(\theta, S^{*(M)}_{1:K}(\theta))$ be the matrix with columns $\nabla_{\theta} g(\mathbf x_{1:K}, \theta_i, \mathbf{\tilde x}_{1:K}^{(m)}))$ for $m=1,\hdots, M$, then
\begin{align} \label{eqn:ddi-alpha}
    \bm{\gamma}^* = -\frac{\nabla_{\theta}g(\theta, S^{*(M)}_{1:K}(\theta))\bm{\alpha}^*}{||\nabla_{\theta}g(\theta, S^{*(M)}_{1:K}(\theta))\bm{\alpha}^*||_2}, \nonumber\\ \bm{\alpha}^* = \argmin_{\bm{\alpha}\in\triangle^M}||\nabla_{\theta}g(\theta, S^{*(M)}_{1:K}(\theta))\bm{\alpha}||^2_2.
\end{align}

\subsection{DDi-AT for NLP classification} \label{sec:ddi-nlp}

The challenge with NLP is that generating strong textual adversarial examples as per Equation \ref{eqn:s-m} can be extremely slow. Hence to increase speed, we generate adversarial examples in the token embedding space, such that we follow Equation \ref{eqn:ddi-alpha}, but adapt Equation \ref{eqn:at-imp-ddi} to,
\begin{align} \label{eqn:ddi-emb}
    g(\mathbf x_{1:K}, \theta_i, \mathbf{\hat{\tilde h}}_{1:K})=\frac{1}{K}\sum_{k} \mathcal L(\mathbf{\hat{\tilde h}}_k, \theta_i), \nonumber \\ \mathbf{\hat{\tilde h}}_k = \argmax_{\mathbf{\tilde h}\in\mathcal S^*(\theta_i, \mathbf h_k)}\left||\nabla_{\theta=\theta_i}\mathcal L(\mathbf{\tilde h}, \theta)\right||_2,
\end{align}
where $\mathbf{h}_k=\{\mathbf h_{k,1}, \hdots, \mathbf h_{k,L}\}$ represents the sequence of token embeddings for tokens $\mathbf{x}_k=\{\mathbf x_{k,1}, \hdots, \mathbf x_{k,L}\}$. We can create our proxy finite set of maximizers, $S^{*(M)}$ (Equation \ref{eqn:s-m}) by using a computer-vision style Projected Gradient Descent (PGD) attack \citep{madry2019deep} in each token embedding space with initialisations of the PGD attack at different points to create multiple adversarial examples,
\begin{equation}
    S^{*(M)}(\theta, \mathbf h) = \{\text{PGD}^{(1)}(\theta, \mathbf h), \hdots, \text{PGD}^{(M)}(\theta, \mathbf h),\}.
\end{equation}
In this work we refer to DDi gradients applied to PGD AT as, \textit{DDi-AT}.

\subsection{Gradient Normalization and Overconfidence} \label{sec:gradnorm}

It is shown in Table \ref{tab:conf-comp} that the use of the DDi gradients with the PGD AT approach (ddi-at) gives rise to a highly overconfident model, which is responsible for the IOR. This section aims to determine the route cause of this overconfidence in the DDi gradient update algorithm. Equation \ref{eqn:at-imp-ddi} indicates that in the DDi gradient update algorithm global gradient normalization is applied. Note that this is different to standard training algorithms where either no normalization is applied or gradient clipping is used where global gradient normalization is only applied if the global gradient norm is larger than a threshold~\citep{DBLP:journals/corr/abs-1211-5063}. Table \ref{tab:conf-comp-ddi} demonstrates that the use of the global gradient normalization in DDi-AT is responsible for the overconfidence and thus IOR. Interestingly, Table \ref{tab:conf-comp-base} reveals that gradient normalization can also induce overconfidence for the standardly trained \textit{baseline} model.

\begin{table}[htb!]
    \centering
    \small
    \begin{tabular}{l|c|cc}
    \toprule
        \textbf{Normalization} & \textbf{clean} &  $\bar P(\hat c|\mathbf{x}_{\text{clean}})$ & $\bar P(\hat c|\mathbf{x}_{\text{adv}})$ \\ \midrule

        gradient norm & $\underset{0.49}{87.90}$ & $\underset{0.03}{99.97}$ & $\underset{0.01}{99.91}$\\
        
        gradient clipping & $\underset{0.68}{88.28}$ & $\underset{0.30}{97.16}$ & $\underset{0.72}{86.12}$\\

        none & $\underset{0.55}{88.20}$ & $\underset{0.42}{96.98}$ & $\underset{0.66}{86.16}$\\
        \bottomrule
    \end{tabular}
    \caption{Model Confidence on clean and adversarial (pwws) examples for DDi-AT model with different forms of gradient normalization in the DDi gradient update step. Rotten Tomatoes dataset, DeBERTa model.}
    \label{tab:conf-comp-ddi}
\end{table}

\begin{table}[htb!]
    \centering
    \small
    \begin{tabular}{l|c|cc}
    \toprule
        \textbf{Normalization} & \textbf{clean} &  $\bar P(\hat c|\mathbf{x}_{\text{clean}})$ & $\bar P(\hat c|\mathbf{x}_{\text{adv}})$ \\ \midrule

        gradient norm & $\underset{0.19}{88.56}$ & $\underset{0.04}{99.96}$ & $\underset{0.02}{99.93}$\\
        
        gradient clipping & $\underset{0.31}{88.94}$ & $\underset{0.29}{97.02}$ & $\underset{0.84}{86.74}$\\

        none & $\underset{0.30}{88.96}$ & $\underset{0.26}{97.08}$ & $\underset{0.68}{86.04}$\\
        \bottomrule
    \end{tabular}
    \caption{Model Confidence on clean and adversarial (pwws) examples for \textit{baseline} model with different forms of gradient normalization in training. Rotten Tomatoes dataset, DeBERTa model.}
    \label{tab:conf-comp-base}
\end{table}

\section{Dataset Descriptions}

We conduct experiments across six standard NLP classification datasets to ensure our findings are robust (statistics summarised in Table \ref{tab:data}). Rotten Tomatoes (\textit{rt}; \citealp{Pang+Lee:05a}) is a binary sentiment classification task for movie reviews. The Emotion Dataset (\textit{emotion}; \citealp{saravia-etal-2018-carer}) categorizes Twitter tweets into one of six emotions: love, joy, surprise, fear, sadness or anger. The remaining three datasets are sourced from the the General Language Understanding Evaluation (GLUE) benchmark~\citep{wang2019glue}.\footnote{For datasets where the provided test set is not labeled, we used the validation set.} The Corpus of Linguistic Acceptability (\textit{cola}) dataset comprises English acceptability judgments sourced from books and journal articles on linguistic theory. Each instance consists of a word sequence annotated to indicate if it is grammatically correct. The Question-answering NLI (\textit{qnli}) dataset assesses the task of sentence pair classification, where one sentence is a question and the other a context. The goal is to ascertain whether the context sentence contains the answer to the question. The Microsoft Research Paraphrase Corpus (\textit{mrpc}) consists of pairs of sentences automatically extracted from online news sources. Human annotations identify if the sentences in each pair are semantically equivalent. Finally we consider the popular AGNews dataset~\citep{Zhang2015CharacterlevelCN}, consisting of articles from 2000 news sources classified into one of four topics: business, sci/tech, world or sports. There are a combined 120,000 training samples and 7600 test samples.

 \begin{table}[htb!]
    \centering
    \small
    \begin{tabular}{l|cccc}
    \toprule
        Dataset & \#classes & Train & Validation & Test \\ \midrule
        rt & 2 & 8.53k & 1.07k & 1.07k\\
        emotion & 6 & 16k & 2k & 2k \\
        cola & 2 & 8.55k & 1.04k & 1.06k\\
        qnli & 2 & 105k & 5.46k & 5.46k \\
        mrpc & 2 & 3.67k & 408 & 1.73k \\
        agnews & 4 & 96k & 24k & 7.6k \\
        \bottomrule
    \end{tabular}
    \caption{Dataset statistics}
    \label{tab:data}
\end{table}

\section{Generative LLMs} \label{sec:llms}

With the advent of powerful generative Large Language Models, their use has become increasingly ubiquitous. However, we found that such popular generative models were not suitable in this work for the following reasons:
\begin{enumerate}
    \item The encoder-only models used in this work are state-of-the-art when fine-tuned on classification tasks.  Some recent studies~\citep{periti2024chatgpt, zhong2023chatgpt} have also found that often (fine-tuned) encoder-only models can be better than the popular generative LLMs for specific classification tasks. We show a comparison of performance in Table \ref{tab:llms-ours} below to a SOTA generative LLM (with zero-shot and few-shot prompting). We also present a comparison of the performance from \citet{zhong2023chatgpt} between BERT-based encoder models and ChatGPT (GPT3.5) on some of the datasets covered in our work (Table \ref{tab:llm-ext}).

    \item As a consequence of the competitive if not superior performance of finetuned encoder-based models on classification tasks, in many industry applications of NLP, such models (BERT-based models) are used extensively due to being light-weight (far fewer parameters than popular generative LLMs), cost-effective and still able to perform extremely well at many classification tasks.

    \item The adversarial attack and defence literature that we are contributing to focuses on encoder models. Therefore, matching their setting allows us to build upon existing attacks and defences.
\end{enumerate}

\begin{table}[htb!]
    \centering
    \small
    \fontsize{6}{9}\selectfont
    \begin{tabular}{p{1.5cm}|c|ccccc}
    \toprule
        Model & \# & rt & agnews & cola & mrpc & qnli\\  \midrule
        DeBERTa-base & 110M & 88.96 & 93.75 & 83.70 & 87.46 & 93.17 \\ \midrule
        Mistral-7B-instruct-0.2 (0-shot)& 7B & 86.47 & 92.32 & 60.25 & 67.15 & 83.11 \\ \midrule
        Mistral-7B-instruct-0.2 (5-shot) & 7B & 88.92 & 94.01 & 78.57 & 76.21 & 89.24 \\
        \bottomrule
    \end{tabular}
    \caption{Comparison of model performance of DeBERTa (used in this paper) with a popular generative LLM, Mistral~\citep{jiang2023mistral}. \# is the number of model parameters.}
    \label{tab:llms-ours}
\end{table}

\begin{table}[htb!]
    \centering
    \small
    \begin{tabular}{l|c|ccc}
    \toprule
         Model & \# & cola & mrpc & qnli  \\ \midrule
        BERT-base &  110M & 56.4 & 90.0 & 84.0 \\
        RoBERTa-base & 110M & 61.8 & 90.0 & 92.0\\ \midrule
        ChatGPT (0 shot) & unk & 56.0 & 66.0 & 84.0 \\
        ChatGPT (1 shot) & unk & 52.0 & 66.0 & 84.0\\
        ChatGPT (5 shot) & unk & 60.2 & 76.0 & 88.0\\
        ChatGPT-CoT & unk & 64.5 & 78.0 & 86.0 \\
         \bottomrule
    \end{tabular}
    \caption{Comparison of encoder-only models and generative LLMs as given in \citet{zhong2023chatgpt}. \# is the number of model parameters.}
    \label{tab:llm-ext}
\end{table}

\iffalse

\begin{table}[htb!]
    \centering
    \small
    \begin{tabular}{lcc}
    \toprule
        Model &  Rotten Tomatoes & AGNews \\ \midrule
        DeBERTa & 88.96 & 93.75 \\
        GPT4 & 89.45 & 94.22 \\
        Mistral-7b-chat & 86.23 & 92.50 \\
        Llama-7b-chat & 85.37 & 91.29 \\
        \bottomrule
    \end{tabular}
    \caption{Comparison of model performance of DeBERTa (used in this paper) with popular generative LLMs. The Generative LLMs are evaluated zero-shot.}
    \label{tab:llms}
\end{table}
\fi

\section{Hyperparameter selection} \label{sec:hyperparam}

We train the Transformer \textit{baseline} models using standard hyper-parameter settings~\citep{DBLP:journals/corr/abs-2006-03654}: initial learning rate of $1e-5$; batch size of 8; total of 5 epochs; 0 warm-up steps;\footnote{We follow TextDefender~\citep{li2021searching} in setting no warm-up steps. Further, empirically validation accuracy remained the same with warm-up of 50 and 100 steps.} ADAMW optimizer, with a weight decay of 0.01 and parameters $\beta_1=0.9$, $\beta_2=0.999$, $\epsilon=1e-8$.

The Adversarial Training (AT) baseline approaches are trained with the same hyperparameters as for the \textit{baseline} model and AT specific hyperparameters are as described in \citet{li-etal-2021-searching}. The default hyperparameters for each baseline (pgd, ascc and freelb) are: 5 adversarial iterations; adversarial learning rate of 0.03; adversarial initialisation magnitude of 0.05; adversarial maximum norm of 1.0; adversarial norm type of l2; $\alpha$ for ascc is 10.0; and $\beta$ for ascc is 40.0. For DDi-AT, DDi gradients are applied to the PGD AT approach, with $M=3$ gradients and $K=3$ PGD iteration steps.

\subsection{DDi-AT Ablation}

The main results report DDi-AT results for DDi gradients applied to PGD AT with $K=3$ PGD steps to find each adversarial example (in the embedding space) during training and $M=3$ adversarial examples (refer to Section \ref{sec:ddi-nlp}). Table \ref{tab:ddi-ablation} gives the impact on adversarial accuracy (with and with out adversarial temperature calibration) of varying $K$ and $M$. It appears that with greater iteration steps, $K$, the model presents a smaller IOR and a greater true robustness as the robustness accuracy does not degrade as much after calibration.

\begin{table}[htb!]
    \small
    \centering
    \begin{tabular}{lll|ccccc}
    \toprule
        \textbf{$M$} & \textbf{$K$} &\textbf{ Adv} & \textbf{clean} & \textbf{pwws} & \textbf{dg}\\ \midrule
        3 & 3 & - & $\underset{\pm0.49}{87.90}$ &   $\underset{\pm0.99}{61.07}$ &  $\underset{\pm1.01}{66.73}$\\
        & & cal & $\underset{\pm0.49}{87.90}$ &   $\underset{\pm1.96}{23.08}$ &  $\underset{\pm3.38}{22.89}$\\ \midrule
        
        3 & 5 & - & $\underset{\pm0.57}{87.87}$ &  $\underset{\pm10.10}{55.53}$ & $\underset{\pm10.06}{61.73}$\\
        & & cal & $\underset{\pm0.57}{87.87}$ &  $\underset{\pm4.61}{31.08}$ & $\underset{\pm6.31}{32.90}$\\ \midrule
        
        3 & 7 & - & $\underset{\pm0.11}{88.12}$ &  $\underset{\pm12.24}{40.06}$ & $\underset{\pm15.79}{44.50}$\\
        & & cal & $\underset{\pm0.11}{88.12}$ &  $\underset{\pm1.26}{31.21}$ & $\underset{\pm0.61}{30.93}$\\ \midrule
        
        % 5 & 3 & - & $\underset{\pm}{}$ & $\underset{\pm}{}$ & $\underset{\pm}{}$ & $\underset{\pm}{}$ & $\underset{\pm}{}$\\
        % & & cal & $\underset{\pm}{}$ & $\underset{\pm}{}$ & $\underset{\pm}{}$ & $\underset{\pm}{}$ & $\underset{\pm}{}$\\ \midrule
        
        5 & 5 & - & $\underset{\pm1.17}{87.65}$ &  $\underset{\pm21.23}{50.59}$ & $\underset{\pm26.22}{54.00}$\\
        & & cal & $\underset{\pm1.17}{87.65}$ &  $\underset{\pm2.05}{28.08}$ & $\underset{\pm4.29}{27.95}$\\ \midrule
        
        5 & 7 & - & $\underset{\pm0.38}{88.15}$ & $\underset{\pm2.96}{31.68}$ &  $\underset{\pm4.79}{34.96}$\\
        & & cal & $\underset{\pm0.38}{88.15}$ & $\underset{\pm1.17}{29.92}$ &  $\underset{\pm0.84}{31.61}$\\
        \bottomrule
    \end{tabular}
    \caption{Ablation: DDi-AT with $M$ PGD adversarial examples, with each PGD adversarial example search during training using $K$ iteration steps.}
    \label{tab:ddi-ablation}
\end{table}

\newpage

\section{Further Experiments for Illusion of Robustness and Mitigation} \label{sec:app-ior}

\subsection{Experiments on Other Datasets} \label{sec:other-data}

Equivalent results are presented for Twitter Emotions (6 emotion classes) in Table \ref{tab:twitter} and for the AGNews dataset (4 news classes) in Table \ref{tab:agnews}.

\begin{table}[htb!]
    \centering
    \small
    \fontsize{8}{11}\selectfont
    \begin{tabular}{l|c|cccc}
    \toprule
       \textbf{Method}  & \textbf{clean} &\textbf{ bae} & \textbf{tf} & \textbf{pwws} & \textbf{dg} \\ \midrule
       
        baseline  &  $\underset{\pm0.24}{93.13}$ &$\underset{\pm0.85}{30.17}$&$\underset{\pm0.55}{5.77}$&$\underset{\pm2.01}{11.80}$&$\underset{\pm2.98}{8.32}$\\ \midrule
        
        $\downarrow$conf (\S \ref{sec:explicit}) & $\underset{\pm0.24}{93.13}$ &$\underset{\pm0.80}{29.63}$&$\underset{\pm0.58}{6.78}$&$\underset{\pm1.55}{15.22}$&$\underset{\pm3.01}{14.68}$\\

         $\uparrow$conf (\S \ref{sec:explicit}) & $\underset{\pm0.24}{93.13}$ &$\underset{\pm0.76}{30.62}$&$\underset{\pm0.51}{16.62}$&$\underset{\pm1.01}{28.85}$&$\underset{\pm2.07}{31.03}$\\
        \midrule
        
        ddi-at (\S\ref{sec:implicit}) & $\underset{\pm0.18}{93.40}$ &$\underset{\pm1.23}{27.92}$&$\underset{\pm0.79}{9.90}$&$\underset{\pm0.67}{18.57}$&$\underset{\pm1.65}{18.17}$\\ \midrule
        
        dg-aug & $\underset{\pm0.11}{92.58}$ &$\underset{\pm2.82}{31.52}$&$\underset{\pm0.25}{4.68}$&$\underset{\pm0.11}{9.33}$&$\underset{\pm0.64}{29.45}$  \\
        
        pgd & $\underset{\pm0.03}{93.48}$ &$\underset{\pm0.43}{28.83}$&$\underset{\pm1.24}{4.88}$&$\underset{\pm0.69}{9.95}$&$\underset{\pm1.08}{5.45}$\\
        
        ascc & $\underset{\pm0.57}{91.15}$ &$\underset{\pm0.23}{34.65}$&$\underset{\pm1.05}{4.60}$&$\underset{\pm0.22}{12.15}$&$\underset{\pm1.40}{11.28}$\\
        
        freelb & $\underset{\pm0.23}{93.67}$ &$\underset{\pm1.00}{29.15}$&$\underset{\pm1.25}{4.93}$&$\underset{\pm0.30}{10.15}$&$\underset{\pm0.73}{5.48}$\\
        
        \bottomrule
    \end{tabular}
    \caption{\textbf{Twitter:} Extreme confidence systems compared to standard AT methods on out-of-the-box adversarial attacks.}
    \label{tab:twitter}
\end{table}

\begin{table}[htb!]
    \centering
    \small
    \fontsize{8}{11}\selectfont
    \begin{tabular}{l|c|cccc}
    \toprule
       \textbf{Method}  & \textbf{clean} & \textbf{bae} & \textbf{tf} &\textbf{ pwws} & \textbf{dg} \\ \midrule
       
        baseline  &  $\underset{\pm0.25}{93.75}$ &$\underset{\pm0.51}{78.46}$&$\underset{\pm1.11}{31.63}$&$\underset{\pm2.93}{42.25}$&$\underset{\pm1.31}{46.21}$\\ \midrule
        
        $\downarrow$conf (\S \ref{sec:explicit}) & $\underset{\pm0.25}{93.75}$ &$\underset{\pm0.51}{81.08}$&$\underset{\pm0.19}{59.17}$&$\underset{\pm2.24}{70.79}$&$\underset{\pm1.06}{75.71}$\\

         $\uparrow$conf (\S \ref{sec:explicit}) & $\underset{\pm0.25}{93.75}$ &$\underset{\pm0.80}{85.71}$&$\underset{\pm0.89}{84.79}$&$\underset{\pm0.36}{88.21}$&$\underset{\pm0.31}{88.17}$\\
        \midrule
        
        ddi-at (\S\ref{sec:implicit}) & $\underset{\pm0.33}{94.25}$ &$\underset{\pm0.75}{88.00}$&$\underset{\pm1.00}{88.08}$&$\underset{\pm0.36}{88.96}$&$\underset{\pm0.13}{89.25}$\\ \midrule
        
        dg-aug & $\underset{\pm0.43}{94.13}$ &$\underset{\pm1.63}{74.58}$&$\underset{\pm0.19}{33.92}$&$\underset{\pm1.25}{50.33}$&$\underset{\pm0.38}{56.38}$  \\
        
        pgd & $\underset{\pm0.50}{94.00}$ &$\underset{\pm0.50}{85.13}$&$\underset{\pm1.27}{45.86}$&$\underset{\pm0.95}{59.58}$&$\underset{\pm1.44}{57.00}$\\
        
        ascc & $\underset{\pm0.46}{94.03}$ &$\underset{\pm0.87}{83.19}$&$\underset{\pm1.95}{49.80}$&$\underset{\pm1.86}{54.04}$&$\underset{\pm1.32}{58.70}$\\
        
        freelb & $\underset{\pm0.07}{93.58}$ &$\underset{\pm0.71}{83.46}$&$\underset{\pm0.66}{44.13}$&$\underset{\pm1.73}{58.13}$&$\underset{\pm2.05}{54.25}$\\
        
        \bottomrule
    \end{tabular}
    \caption{\textbf{AGNews:} Extreme confidence systems compared to standard AT methods on out-of-the-box adversarial attacks. \textit{*Evaluation on 1000 samples.}}
    \label{tab:agnews}
\end{table}

\subsection{Experiments on Other Models} \label{sec:other-models}

The \textit{illusion of robustness} is presented for an overconfident, underconfident and DDi-AT \textit{DeBERTa} model in the main paper in Table \ref{tab:attack}. The same trends are observed for other popular Transformer-encoder (\textit{base}) models: RoBERTa (Table \ref{tab:roberta-ior}); and BERT (Table \ref{tab:bert-ior}).

\begin{table}[htb!]
    \centering
    \small
    \begin{tabular}{l|c|cccc}
    \toprule
       \textbf{Method}  & \textbf{clean} & \textbf{bae} &\textbf{ tf} & \textbf{pwws} & \textbf{dg} \\ \midrule
       
        baseline  & $\underset{\pm0.47}{88.27}$ & $\underset{\pm0.74}{32.46}$ & $\underset{\pm0.72}{17.01}$ & $\underset{\pm0.05}{21.23}$ & $\underset{\pm1.71}{24.30}$ \\ \midrule
        
        $\downarrow$conf & $\underset{\pm0.47}{88.27}$ & $\underset{\pm0.33}{31.77}$ & $\underset{\pm1.27}{20.42}$ & $\underset{\pm1.43}{24.92}$ & $\underset{\pm1.33}{32.99}$\\

         $\uparrow$conf & $\underset{\pm0.47}{88.27}$ & $\underset{\pm0.76}{37.65}$ & $\underset{\pm0.94}{53.63}$ & $\underset{\pm0.61}{58.66}$ & $\underset{\pm0.92}{66.32}$\\
        \midrule
        
        ddi-at & $\underset{\pm0.62}{88.06}$ & $\underset{\pm0.85}{36.24}$ & $\underset{\pm0.41}{50.84}$ & $\underset{\pm1.25}{54.85}$ & $\underset{\pm1.27}{62.76}$\\
        
        \bottomrule
    \end{tabular}
    \caption{\textbf{RoBERTa} Model: Robustness of Mis-calibrated systems.}
    \label{tab:roberta-ior}
\end{table}

% \begin{table}[htb!]
%     \centering
%     \small
%     \begin{tabular}{l|c|cccc}
%     \toprule
%       Method  & clean & bae & tf & pwws & dg \\ \midrule
       
%         base  & $\underset{\pm0.71}{89.81}$ & $\underset{\pm0.92}{34.05}$ & $\underset{\pm2.42}{24.23}$ & $\underset{\pm1.81}{27.02}$ & $\underset{\pm2.66}{27.14}$ \\ \midrule
        
%         $\downarrow$conf & $\underset{\pm0.71}{89.81}$ & $\underset{\pm1.14}{33.30}$ & $\underset{\pm0.19}{29.55}$ & $\underset{\pm1.19}{33.65}$ & $\underset{\pm0.73}{38.77}$\\

%          $\uparrow$conf & $\underset{\pm0.71}{89.81}$ & $\underset{\pm0.54}{39.65}$ & $\underset{\pm0.76}{57.10}$ & $\underset{\pm0.93}{63.54}$ & $\underset{\pm0.34}{65.85}$\\
%         \midrule
        
%         DDi & $\underset{\pm0.50}{90.34}$ & $\underset{\pm1.28}{41.24}$ & $\underset{\pm2.39}{61.63}$ & $\underset{\pm1.72}{65.48}$ & $\underset{\pm0.32}{67.64}$\\
        
%         \bottomrule
%     \end{tabular}
%     \caption{\textbf{ELECTRA} Model: Robustness of Mis-calibrated systems.}
%     \label{tab:electra}
% \end{table}

\begin{table}[htb!]
    \centering
    \small
    \begin{tabular}{l|c|cccc}
    \toprule
      \textbf{ Method}  & \textbf{clean} &\textbf{ bae }& \textbf{tf} & \textbf{pwws} & \textbf{dg} \\ \midrule
       
        baseline  & $\underset{\pm0.50}{85.08}$ & $\underset{\pm0.76}{30.52}$ & $\underset{\pm0.32}{21.01}$ & $\underset{\pm0.34}{21.20}$ & $\underset{\pm2.14}{23.14}$ \\ \midrule
        
        $\downarrow$conf & $\underset{\pm0.50}{85.08}$ & $\underset{\pm0.19}{29.74}$ & $\underset{\pm0.53}{20.95}$ & $\underset{\pm1.36}{24.58}$ & $\underset{\pm0.24}{30.64}$\\

         $\uparrow$conf & $\underset{\pm0.50}{85.08}$ & $\underset{\pm1.11}{35.08}$ & $\underset{\pm0.85}{45.84}$ & $\underset{\pm1.37}{53.25}$ & $\underset{\pm2.06}{57.50}$\\
        \midrule
        
        ddi-at & $\underset{\pm0.43}{85.55}$ & $\underset{\pm0.29}{36.80}$ & $\underset{\pm0.69}{48.09}$ & $\underset{\pm1.04}{51.50}$ & $\underset{\pm1.16}{56.60}$\\
        
        \bottomrule
    \end{tabular}
    \caption{\textbf{BERT} Model: Robustness of Mis-calibrated systems.}
    \label{tab:bert-ior}
\end{table}

\newpage
\subsection{Transferability Defence against IOR} \label{sec:ior-transfer}

One might even argue that to expose an IOR, it is unnecessary for an adversary to modify the model with adversarial temperature scaling to find adversarial examples. Instead, adversarial examples can be found for another model (e.g., \textit{baseline}) and transferred to the target model. This follows from \citet{DBLP:journals/corr/abs-1809-02861} where it is shown that similar architectures can be susceptible to the same adversarial examples. 
We test this hypothesis. It is clear from Table \ref{tab:transferability} that although the transfer attack from \textit{baseline} to \textit{ddi-at} is effective in reducing the adversarial accuracy, it is unable to bring the adversarial accuracy down to the same values as \textit{baseline}, as achieved by our proposed temperature scaling approaches in Table \ref{tab:cal}.

\begin{table}[t]
    \centering
    \small
    \fontsize{7}{11}\selectfont
    \begin{tabular}{ll|c|cccc}
    \toprule
\textbf{source} & \textbf{target} & \textbf{clean} & \textbf{bae} & \textbf{tf} & \textbf{pwws} & \textbf{dg}\\ \midrule
        baseline & baseline& $\underset{\pm0.30}{88.96}$ & $\underset{\pm1.20}{31.39}$ & $\underset{\pm0.49}{17.82}$ & $\underset{\pm0.62}{20.42}$ & $\underset{\pm0.94}{20.11}$\\
        
        ddi-at & ddi-at & $\underset{\pm0.49}{87.90}$ &  $\underset{\pm0.75}{39.18}$ &  $\underset{\pm1.67}{56.54}$ &  $\underset{\pm0.99}{61.07}$ &  $\underset{\pm1.01}{66.73}$\\ \midrule
        
        baseline & ddi-at & $\underset{\pm0.49}{87.90}$ & $\underset{\pm0.60}{48.91}$ & $\underset{\pm1.15}{52.47}$ & $\underset{\pm1.64}{50.00}$ & $\underset{\pm0.99}{48.53}$\\
        \bottomrule
    \end{tabular}
    \caption{Transferability: adversarial examples for each attack method are generated for the source model and adversarial accuracy (\%) is given for the target model.}
    \label{tab:transferability}
\end{table}

\newpage

\subsection{Calibration Error} \label{sec:cal-error}

In Table \ref{tab:cal-error} we verify that the calibration approaches are effective in calibrating the models. We report the metrics: Expected Calibration Error (ECE) and Maximum Calibration Error (MCE).

\begin{table}[htb!]
    \centering
    \small
    \begin{tabular}{l|cc|cc}
    \toprule
    \textbf{Method} & \textbf{ECE} &\textbf{ MCE}  & $\bar P(\hat c|\mathbf{x}_{\text{clean}})$ & $\bar P(\hat c|\mathbf{x}_{\text{adv
        }})$\\ \midrule
         baseline & $\underset{\pm0.62}{48.82}$ & $\underset{\pm1.15}{51.98}$ & $\underset{\pm0.26}{97.08}$ & $\underset{\pm0.68}{86.04}$\\ \midrule
         
         \textbf{$\downarrow$conf} & $\underset{\pm0.30}{38.96^*}$ & $\underset{\pm0.30}{38.96^*}$ & $\underset{\pm0.00}{50.00007}$ & $\underset{\pm0.00}{50.00004}$\\
         +cal & $\underset{\pm0.30}{38.96^*}$ & $\underset{\pm0.30}{38.96^*}$ & $\underset{\pm0.00}{50.00004}$ & $\underset{\pm0.00}{50.00002}$\\ \midrule
         
        \textbf{$\uparrow$conf} & $\underset{\pm1.03}{51.31}$ & $\underset{\pm11.8}{62.62}$ & $\underset{\pm0.02}{99.98}$ & $\underset{\pm0.01}{99.95}$\\
         +cal & $\underset{\pm0.91}{42.30}$ & $\underset{\pm1.04}{48.28}$ & $\underset{\pm0.45}{90.36}$ & $\underset{\pm0.58}{75.88}$\\ \midrule
         
         \textbf{ddi-at} & $\underset{\pm0.57}{52.41}$ & $\underset{\pm20.97}{74.87}$ & $\underset{\pm0.03}{99.97}$ & $\underset{\pm0.05}{99.91}$\\
         +cal & $\underset{\pm0.58}{42.60}$ & $\underset{\pm18.36}{62.73}$ & $\underset{\pm0.11}{90.13}$ & $\underset{\pm0.80}{87.54}$\\
    \bottomrule
    \end{tabular}
    \caption{Calibration Error and Average Predicted Confidence (on clean and adv-pwws). N.B. baseline is across 3 seeds. *off-the-shelf calibration error computation fails here as all confidences very close to 50\%, so manual computation of CE here: \textit{accuracy - 50\%}.}
    \label{tab:cal-error}
\end{table}

\subsection{IOR in AT Approaches} \label{sec:cal-base}

The main results demonstrate that highly miscalibrated systems have an \textit{illusion of robustness} (IOR), where an adversary's temperature calibration can mitigate this illusion of robustness. Considering the rotten tomatoes dataset and the DeBERTa model, Table \ref{tab:ior-baseline-gradnorm} demonstrates that standard  AT approaches considered in this work can also suffer from the IOR, when global gradient normalization is included in the training algorithm (Note that Table \ref{tab:conf-comp-base} shows that gradient normalization can be a source of model overonfidence). Nevertheless, Table \ref{tab:ior-baseline} demonstrates that when global gradient normalization is excluded from the training algorithm, the baseline AT approaches considered in this work no longer present IORs as calibration does not degrade their adversarial accuracy.

\begin{table}[htb!]
    \centering
    \small
    \fontsize{8}{11}\selectfont
    \begin{tabular}{lc|c|cccc}
    \toprule
      \textbf{ Method} & \textbf{Adv} & \textbf{clean} & \textbf{bae} & \textbf{tf} & \textbf{pwws} & \textbf{dg} \\ \midrule
        pgd$^*$ & - & $\underset{\pm0.64}{88.59}$& $\underset{\pm0.55}{39.94}$ & $\underset{\pm1.04}{58.02}$  & $\underset{\pm0.77}{64.45}$ & $\underset{\pm0.83}{67.02}$ \\
        & cal & $\underset{\pm0.64}{\underline{88.59}}$ &$\underset{\pm0.20}{33.71}$&$\underset{\pm0.86}{17.73}$&$\underset{\pm1.80}{25.20}$&$\underset{\pm1.46}{25.74}$\\ \midrule

        ascc$^*$ & - & $\underset{\pm0.36}{87.77}$ & $\underset{\pm0.69}{40.01}$ & $\underset{\pm1.57}{54.32}$ & $\underset{\pm0.86}{63.99}$ & $\underset{\pm0.93}{67.43}$ \\
        & cal & $\underset{\pm0.36}{87.77}$ &$\underset{\pm0.64}{33.61}$&$\underset{\pm2.17}{15.13}$&$\underset{\pm0.77}{23.50}$&$\underset{\pm2.11}{26.80}$\\
        \bottomrule
    \end{tabular}
    \caption{Baseline AT approach (PGD and ASCC results here) can also suffer from IOR (calibration reduces observed adversarial robustness) when global gradient normalization used in the training algorithm. The IOR was also observed for dg-aug and freelb AT schemes.}
    \label{tab:ior-baseline-gradnorm}
\end{table}

\begin{table}[htb!]
    \centering
    \small
    \fontsize{8}{11}\selectfont
    \begin{tabular}{lc|c|cccc}
    \toprule
      \textbf{ Method} & \textbf{Adv} & \textbf{clean} & \textbf{bae} & \textbf{tf} & \textbf{pwws} & \textbf{dg} \\ \midrule

        baseline &-& $\underset{\pm0.30}{88.96}$ & $\underset{\pm1.20}{31.39}$ & $\underset{\pm0.49}{17.82}$ & $\underset{\pm0.62}{20.42}$ & $\underset{\pm0.94}{20.11}$\\
        &cal& $\underset{\pm0.30}{88.96}$ & $\underset{\pm1.20}{31.39}$ & $\underset{\pm0.51}{17.80}$ & $\underset{\pm0.66}{20.46}$ & $\underset{\pm0.88}{20.05}$\\ \midrule
        
        dg-aug & - & $\underset{\pm0.39}{87.12}$ &$\underset{\pm1.59}{34.74}$&$\underset{\pm1.83}{22.36}$&$\underset{\pm2.57}{26.11}$&$\underset{\pm0.75}{37.43}$\\
        &cal& $\underset{\pm0.39}{87.12}$ & $\underset{\pm1.59}{34.74}$ & $\underset{\pm1.81}{22.36}$ & $\underset{\pm2.32}{25.98}$ & $\underset{\pm0.74}{37.45}$\\ \midrule
        
        pgd & - & $\underset{\pm0.73}{88.24}$ &$\underset{\pm0.57}{33.65}$&$\underset{\pm0.47}{19.92}$&$\underset{\pm0.87}{26.70}$&$\underset{\pm0.61}{26.05}$\\
        &cal& $\underset{\pm0.73}{88.24}$ & $\underset{\pm0.57}{33.65}$ & $\underset{\pm0.46}{19.90}$ & $\underset{\pm0.90}{26.74}$ & $\underset{\pm0.54}{26.10}$\\ \midrule
        
        ascc & - & $\underset{\pm0.36}{87.77}$ &$\underset{\pm0.64}{33.61}$&$\underset{\pm2.17}{15.13}$&$\underset{\pm0.77}{23.50}$&$\underset{\pm2.11}{26.80}$\\
        &cal& $\underset{\pm0.36}{87.77}$ & $\underset{\pm0.63}{33.60}$ & $\underset{\pm2.19}{15.10}$ & $\underset{\pm0.79}{23.49}$ & $\underset{\pm2.03}{26.75}$\\ \midrule

        freelb & - & $\underset{\pm0.32}{88.74}$ &$\underset{\pm0.52}{32.52}$&$\underset{\pm1.70}{19.51}$&$\underset{\pm0.70}{24.55}$&$\underset{\pm0.73}{24.52}$\\
        &cal& $\underset{\pm0.32}{88.74}$ & $\underset{\pm0.32}{88.74}$ & $\underset{\pm1.72}{19.50}$ & $\underset{\pm0.55}{24.35}$ & $\underset{\pm0.75}{24.54}$\\
        \bottomrule
    \end{tabular}
    \caption{Baseline AT approach can be freed of the IOR when global gradient normalization is not used in the training algorithm.}
    \label{tab:ior-baseline}
\end{table}

\subsection{Alternative Calibration Approaches} \label{sec:cal-compare}

In the main results, temperature calibration was implemented to detect adversarial examples based on two central considerations: 1) Temperature calibration effectively facilitates the adversarial attack search, especially for obviously mis-calibrated models; and
2) Temperature calibration preserves the rank order of logits, thereby ensuring transferability of adversarial examples from the calibrated to the original uncalibrated model. To broaden the analytical scope, alternative calibration techniques are examined. The goal is to assess their potential in mitigating the disruption to the adversarial attack search processes and to determine the potency of the resulting adversarial examples on the uncalibrated model. Binning-based calibration is deemed unsuitable due to its intrinsic non-differentiability, which could prevent the adversarial search process. Hence, the multi-class version of Platt Scaling is explored as a viable calibration strategy and subsequently contrasted against the benchmark temperature calibration approach from the main results. The performance of the calibration results is shown in Table \ref{tab:cal-platt}, where it is evident that the Platt scaling approach is far less stable than temperature calibration and can in fact excessively enhance the \textit{illusion of robustness}.

For automatic calibration, standard training hyperparameters were employed. Specifically, the temperature calibration protocol was set at 5,000 iterations with a learning rate of 0.01. Similarly, the Platt scaling protocol was also designed for 5000 iterations with a learning rate of 0.01. A point to note for practical implementation: adversaries might need to refine calibrator hyperparameters to minimize the Expected Calibration Error (ECE) on a specified validation set. However, ECE determination is nuanced, largely due to its sensitivity to chosen bin widths, as highlighted in Table \ref{tab:cal-error} for instances of underconfidence.

\begin{table}[htb!]
    \centering
    \small
    \fontsize{6}{7}\selectfont
    \begin{tabular}{lc|c|cccc}
    \toprule
       \textbf{Method} & \textbf{Adv} & \textbf{clean} & \textbf{bae} & \textbf{tf} & \textbf{pwws} & \textbf{dg} \\ \midrule
        baseline  & -& $\underset{\pm0.30}{88.96}$ & $\underset{\pm1.20}{31.39}$ & $\underset{\pm0.49}{17.82}$ & $\underset{\pm0.62}{20.42}$ & $\underset{\pm0.94}{20.11}$\\ \midrule

        $\downarrow$conf &-& $\underset{\pm0.30}{88.96}$ & $\underset{\pm0.94}{31.21}$ & $\underset{\pm0.99}{20.98}$ & $\underset{\pm0.89}{25.17}$ & $\underset{\pm2.78}{32.18}$\\
        &temp& $\underset{\pm0.30}{88.96}$ & $\underset{\pm0.34}{31.52}$ & $\underset{\pm0.43}{21.89}$ & $\underset{\pm1.31}{27.58}$ & $\underset{\pm0.34}{31.52}$\\
        &platt & $\underset{\pm0.30}{88.96}$ & $\underset{\pm12.15}{72.08}$ & $\underset{\pm18.00}{70.33}$ & $\underset{\pm16.72}{72.70}$ & $\underset{\pm17.11}{74.73}$\\ \midrule 
        
        $\uparrow$conf &-& $\underset{\pm0.30}{88.96}$ & $\underset{\pm1.18}{37.71}$ & $\underset{\pm0.73}{54.35}$ & $\underset{\pm0.62}{59.29}$ & $\underset{\pm1.81}{65.60}$\\
        &temp& $\underset{\pm0.30}{88.96}$ & $\underset{\pm1.20}{31.39}$ & $\underset{\pm0.49}{17.82}$ & $\underset{\pm0.74}{20.45}$ & $\underset{\pm1.46}{21.64}$\\
        &platt &  $\underset{\pm0.30}{88.96}$ & $\underset{\pm3.73}{37.21}$ & $\underset{\pm17.90}{34.55}$ & $\underset{\pm19.70}{37.46}$ & $\underset{\pm19.59}{41.09}$\\ \midrule

        ddi-at &-& $\underset{\pm0.49}{87.90}$ &  $\underset{\pm0.75}{39.18}$ &  $\underset{\pm1.67}{56.54}$ &  $\underset{\pm0.99}{61.07}$ &  $\underset{\pm1.01}{66.73}$\\
        &temp& $\underset{\pm0.49}{87.90}$ &  $\underset{\pm0.57}{31.80}$ &  $\underset{\pm3.01}{18.36}$ &  $\underset{\pm1.96}{23.08}$ &  $\underset{\pm3.38}{22.89}$\\
        &platt & $\underset{\pm0.49}{87.90}$ & $\underset{\pm19.42}{43.34}$ & $\underset{\pm32.23}{38.77}$ & $\underset{\pm31.66}{42.25}$ & $\underset{\pm32.72}{42.72}$ \\
        \bottomrule
    \end{tabular}
    \caption{Adversarial mitigation of highly miscalibrated systems using different test-time calibration approaches.}
    \label{tab:cal-platt}
\end{table}

\subsection{Extreme Miscalibration Leads to Masked Gradients} \label{sec:noisy-grad}

Section \ref{sec:conf} argues that for heavily miscalibrated systems, the `gradients' of the output probabilities with respect to the input are extremely noisy. Therefore, of-the-shelf adversarial attack methods, that use these gradients to select which tokens in the input sequence to attack, receive noisy signals and fail to operate. In this section, we demonstrate that extreme miscalibration does indeed cause noisy gradients for off-the-shelf-adversarial attacks. Note that these noisy gradients are referred to as \textit{obfuscated} gradients or gradient masking by \citet{DBLP:journals/corr/abs-1802-00420}.

We consider two systems: the standard \textit{baseline} system from the main paper and the heavily miscalibrated, overconfident system, $\uparrow$conf in the main paper. Experiments are on the \textit{rt} dataset and we consider specifically the PWWS attack and Textfooler attack. These off-the-shelf adversarial attack approach rank all tokens $w_i$ in the input sequence $\mathbf x$ by their influence on the output of the model (N.B. this is considered an approximation for the gradient of the output with respect to each input token). The PWWS attack refers to this influence as \textit{saliency}, whilst the Textfooler attack calls it \textit{importance}. To assess the impact of heavy miscalibration on the rank ordering, Table \ref{tab:spearman} reports the Spearman Rank Correlation between the rank of all input tokens (in the first iteration of the attack) as per the two models: \textit{baseline} and $\uparrow$conf. The average correlation and standard deviation are given over the entire dataset. The average rank correlation is 0.28 for PWWS and 0.29 for Textfooler, which is very low and demonstrates that by simply having heavy miscalibration there is a significant impact on the attack mechanism. Further, the standard deviation is also large, suggesting that for many input sequences, the correlation is even lower. 
\begin{table}[htb!]
    \centering
    \small
    \begin{tabular}{lc}
    \toprule
        Attack & Rank Correlation \\ \midrule
         pwws & $\underset{\pm0.24}{0.28}$\\
         textfooler & $\underset{\pm 0.26}{0.29}$\\
         \bottomrule
    \end{tabular}
    \caption{Spearman Rank Correlation of input tokens' importance with (overonfident model) and without (\textit{baseline} model) heavy miscalibration. The low rank correlation demonstrates that the token importance is strongly impacted by extreme confidence, which can explain the observed IOR for highly miscalibrated models.}
    \label{tab:spearman}
 
\end{table}

\newpage

\section{Further Experimental Results for High Temperature Training for Genuine Robustness}

Here ST$^*$ will refer to a standardly trained \textit{baseline} model (Equation \ref{eqn:standard}), with gradient normalization during training whilst AT will refer to an adversarially trained model (Equation \ref{eqn:robust}), as indicated by the naming convention given in Table \ref{tab:naming}.

\begin{table}[htb!]
    \centering
    \small
    \begin{tabular}{l|cc}
         & standard & adversarial \\ \midrule
         $T=1$ & ST & AT \\
         High $T$ & ST $\oplus \ T$ & AT $\oplus \ T$\\
    \end{tabular}
    \caption{Naming convention for experiments with different training objectives and high temperature training.}
    \label{tab:naming}
\end{table}

\subsection{Class Margin Explanation}

\begin{figure}[htb!]
     \centering
     \begin{subfigure}[b]{0.45\linewidth}
         \centering
         \includegraphics[width=\columnwidth]{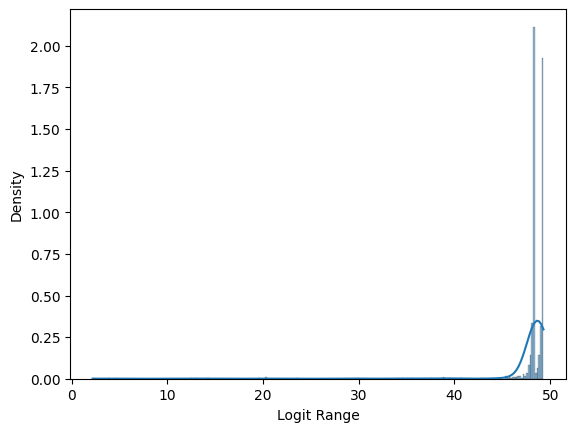}
         \caption{$T=1$}
     \end{subfigure}
     \hfill
     \begin{subfigure}[b]{0.45\linewidth}
         \centering
         \includegraphics[width=\columnwidth]{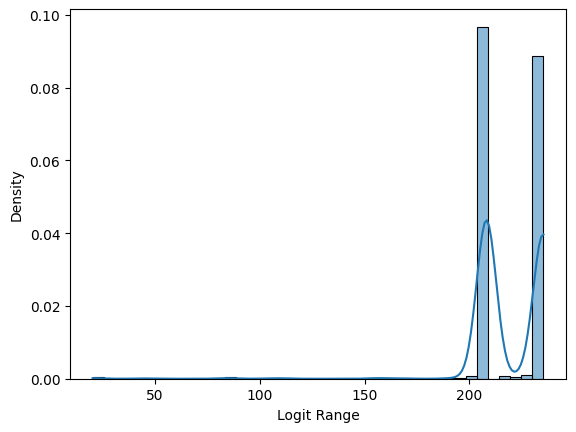}
         \caption{$T=100$}
     \end{subfigure}
        \caption{Probability Density (histogram plot) of predicted class logits' range (smallest logit subtracted from largest logit value) on \textit{rt} test set with and without a high training temperature for the \textit{baseline} ST DeBERTa model. The higher temperature training setting ($T=100$) has a larger class logits' range, suggesting that an adversarial attack has to make a greater change in the logit space to be successful in changing the predicted class.}
        \label{fig:logits}
\end{figure}

The success of high temperature training for adversarial robustness can perhaps be explained by considering the size of the class margin~\citep{robey2023adversarial}. A high temperature smooths the probability distribution across classes, such that the probabilities of the different classes are closer together. To minimize the cross entropy loss during the training, the model's parameters learn to compensate for this smoothing by pushing the logits of the different classes further apart (we see this in Figure \ref{fig:logits}, where the range of logits substantially increases with higher training temperatures). Intuitively, this can be viewed as increasing the distance to the class boundary in the logit space and thus making it more difficult for an adversarial attack to change the predicted class, giving rise to the observed increase in adversarial robustness. Future work will aim to rigorously understand and explain the observed robustness gains of training with a high temperature.

\subsection{High Temperature Standard Training} \label{sec:ht-results}

\begin{figure*}[htb!]
     \centering
     \begin{subfigure}[b]{0.45\linewidth}
         \centering
         \includegraphics[width=\columnwidth]{Figures/temp/rt.png}
         \caption{rotten tomatoes}
     \end{subfigure}
     \hfill
     \begin{subfigure}[b]{0.45\linewidth}
         \centering
         \includegraphics[width=\columnwidth]{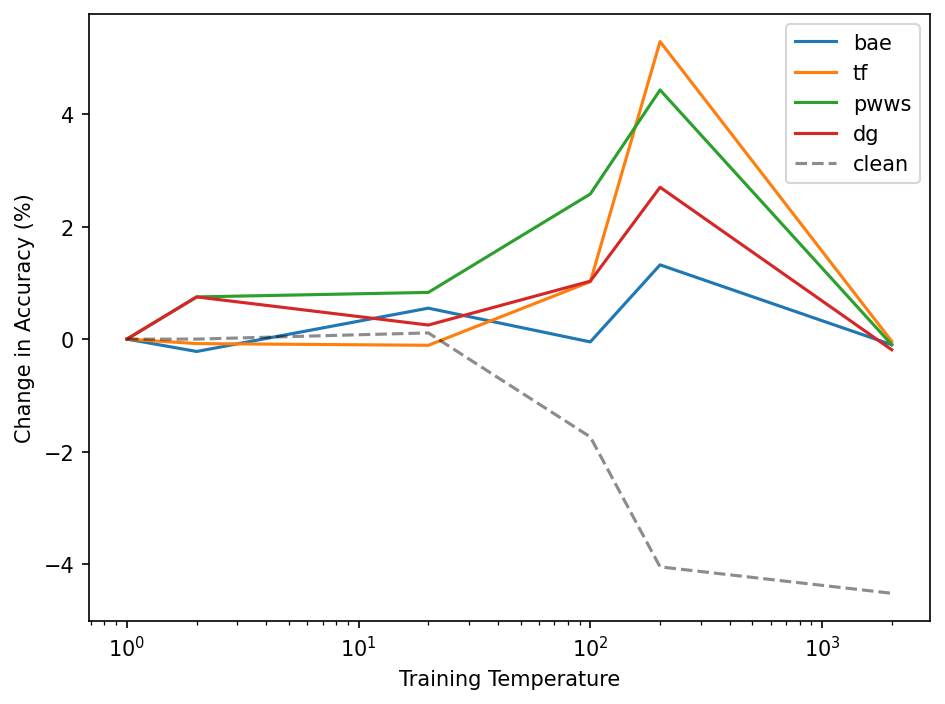}
         \caption{emotion}
     \end{subfigure}
     \hfill
     \begin{subfigure}[b]{0.45\linewidth}
         \centering
         \includegraphics[width=\columnwidth]{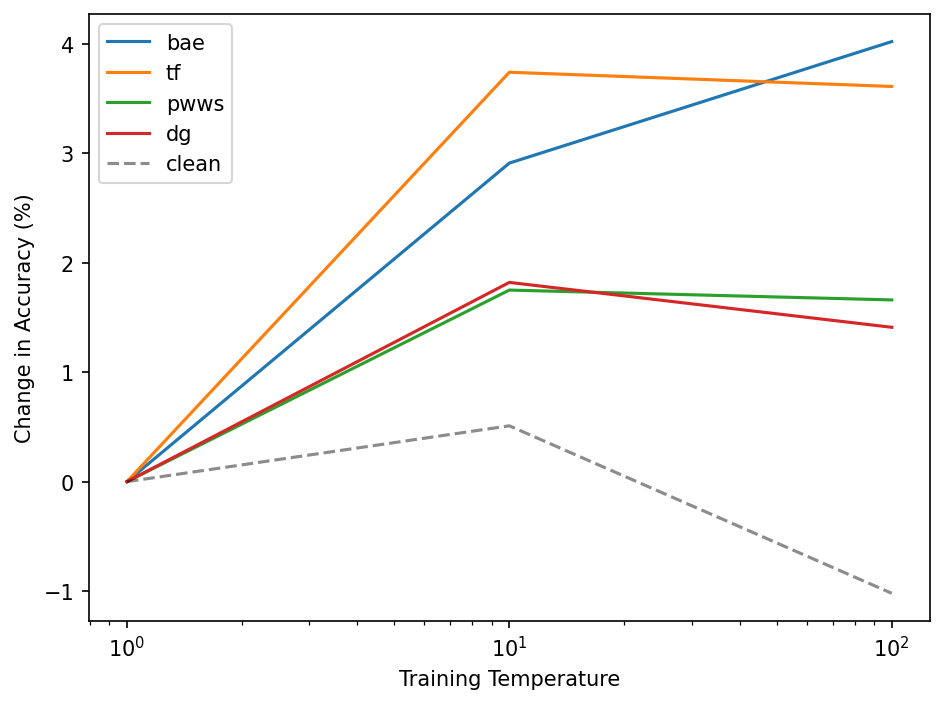}
         \caption{cola}
     \end{subfigure}
     \hfill
     \begin{subfigure}[b]{0.45\linewidth}
         \centering
         \includegraphics[width=\columnwidth]{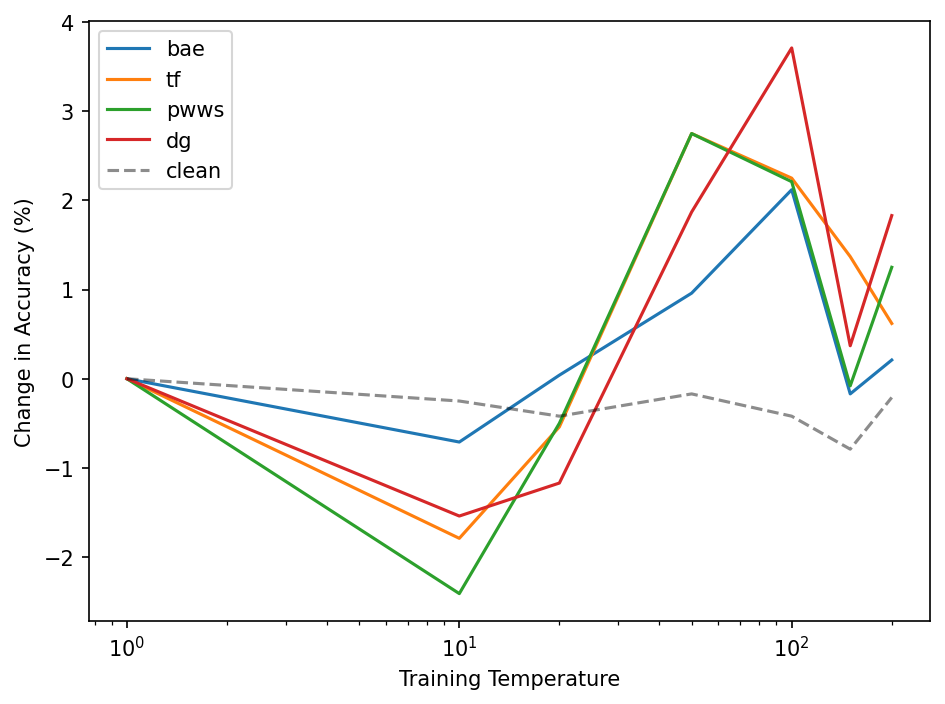}
         \caption{qnli}
     \end{subfigure}
\hfill
     \begin{subfigure}[b]{0.45\linewidth}
         \centering
         \includegraphics[width=\columnwidth]{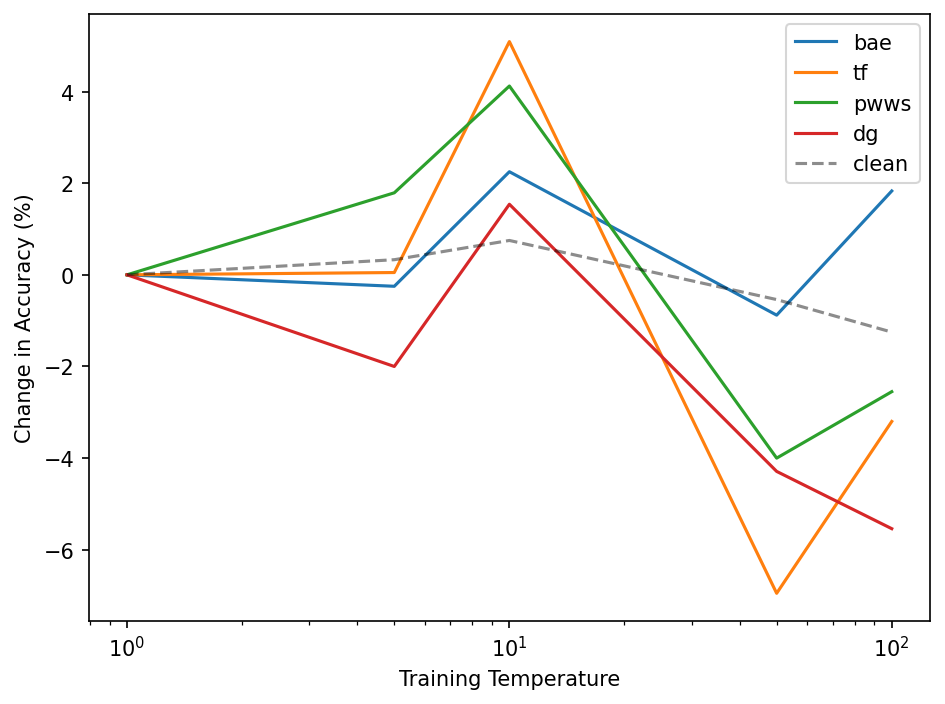}
         \caption{mrpc}
     \end{subfigure}
        \caption{The use of a training temperature, $T$, is a simple adjustment in standard model training (ST), where the temperature parameter, $T$, is used to scale down predicted model logits. Higher training temperatures enhance model robustness against unseen adversarial attacks (\textit{bae, tf, pwws, dg}) without requiring prior knowledge of these attack forms during training. This increased robustness is quantified by the absolute change in adversarial accuracy compared to the baseline $T=1$ ST model.}
        \label{fig:temp}
\end{figure*}

For each dataset, Figure \ref{fig:temp} presents the change in clean and adversarial accuracy of a standardly trained baseline \textit{ST} model trained as per a standard training objective (Equation \ref{eqn:standard}), with different temperatures $T$ used during training. We present the detailed breakdown of the clean and adversarial accuracies for each training temperature, for each adversarial attack and each dataset, in Appendix \ref{app:heat-detailed-datasets}.

We first observe a general increase in adversarial accuracy (robustness) with the training temperature and then a decrease in the accuracy with extremely large training temperatures,\footnote{The drop in robustness for extremely large training temperatures may be attributed to large temperatures excessively smoothing the predicted probability distribution during training, which makes it too challenging for the model to learn, as is reflected by the significant decrease in clean accuracy.} across all datasets. In some datasets (e.g., qnli and mrpc) there is a slight decrease in robustness before the sudden rise in robustness. Nevertheless, there exists a consistent robustness profile for each dataset, where robustness peaks at similar temperatures for all tested adversarial attack types (\textit{bae, tf, pwws, and dg}). This is particularly useful, as a model developer, with access to only one form of adversarial attack, can tune the training temperature for optimal robustness on that specific attack form, yet be confident that the robustness gains will also transfer to the other unseen/unknown attack forms.

A further observation is that increasing temperature can lead to a small drop (between 1\% and 4\%) in clean accuracy. This is perhaps expected as the model can be viewed as being trained in a mode further from the optimal hyper-parameter setting. However, across all the datasets, the optimal temperature (aligned with the peak in adversarial accuracy) results in a maximal drop in clean accuracy of 1\% (apart from for the emotion dataset). Given the gains in adversarial accuracy can be between 4\% and 14\%, this trade-off for clean accuracy can be acceptable. Further, a model developer can choose to operate at a different operating point, by selecting a training temperature that gives a smaller drop in clean accuracy (and settle for a less significant gain in the model robustness).

~\newpage
~\newpage
\newpage

\subsection{High Temperature Adversarial Training} \label{sec:ht-at}

Here we explore the impact of combining the high temperature training approach with popular NLP AT methods. We consider four popular adversarial training approaches: dg-aug$^*$, PGD$^*$, FreeLB$^*$ and ASCC$^*$. Table \ref{tab:AT-gradnorm-rt-app} and Table \ref{tab:AT-gradnorm-twitter} give the baseline ST$^*$ results and AT results combined with the temperature training approach on the \textit{rt} and \textit{emotion} datasets, respectively. 

Although more significant for \textit{rt} than \textit{emotion}, for both datasets, combining with the high training temperature approach improves the adversarial accuracy for all adversarial attack forms (\textit{bae}, \textit{tf}, \textit{pwws}, and \textit{dg}) for the different adversarial training approaches PGD$^*$, FreeLB$^*$, and ASCC$^*$. This demonstrates that high temperature training is complementary with such adversarial training approaches and thus consistently encourages a gain in robustness. Interestingly, we observe that for \textit{dg-aug}, high temperature training is able to consistently improve adversarial accuracy for \textit{bae}, \textit{tf}, and \textit{pwws} adversarial attacks, but can cause a drop in adversarial accuracy for the \textit{dg} attack. It should be emphasized that \textit{dg} in this context behaves as a \textit{seen} attack form, as the training uses augmentation with \textit{dg} adversarial examples, whilst the other attacks (\textit{bae}, \textit{tf}, and \textit{pwws}) can be considered \textit{unseen} attack forms that the model developer has no knowledge of during training. This suggests that for augmentation-based NLP adversarial training approaches, a high training temperature does not necessarily increase robustness to \textit{seen} attack forms, but is successful in boosting robustness to \textit{unseen} attack forms.

\begin{table}[t]
    \centering
    \small
    \begin{tabular}{l|c|cccc}
    \toprule
        Method & clean & bae & tf & pwws & dg \\ \midrule
        
        \textbf{baseline}$^*$ &$\underset{\pm0.19}{88.56}$ & $\underset{\pm0.14}{32.40}$ & $\underset{\pm0.48}{18.79}$ & $\underset{\pm1.22}{21.36}$ & $\underset{\pm0.66}{21.11}$  \\

         $\oplus\ T$ &$\underset{\pm0.44}{87.55}$ & $\underset{\pm0.84}{\underline{35.83}}$ & $\underset{\pm4.57}{\underline{26.83}}$ & $\underset{\pm3.07}{\underline{31.49}}$ & $\underset{\pm4.71}{\underline{35.18}}$  \\ \midrule

       % \textbf{ pwws-aug} &$\underset{\pm0.86}{87.24}$ &$\underset{\pm0.76}{35.26}$&$\underset{\pm0.95}{23.09}$&$\underset{\pm3.02}{35.70}$&$\underset{\pm0.88}{33.86}$  \\
       %   $\oplus$gT & $\underset{\pm0.34}{86.02}$ & $\underset{\pm0.38}{\textbf{37.27}}$ & $\underset{\pm0.44}{\textbf{31.49}}$ & $\underset{\pm0.81}{\textbf{40.84}}$ & $\underset{\pm0.62}{\textbf{39.81}}$ \\\midrule\midrule
        
        \textbf{pgd}$^*$&$\underset{\pm0.64}{\underline{88.59}}$ &$\underset{\pm0.20}{33.71}$&$\underset{\pm0.86}{17.73}$&$\underset{\pm1.80}{25.20}$&$\underset{\pm1.46}{25.74}$  \\
        $\oplus\ T$ & $\underset{\pm0.43}{87.77}$& $\underset{\pm0.33}{\underline{34.77}}$& $\underset{\pm1.76}{\underline{24.55}}$& $\underset{\pm1.08}{\underline{31.46}}$& $\underset{\pm2.64}{\underline{31.77}}$\\ \midrule
        
        \textbf{freelb}$^*$ &$\underset{\pm0.32}{\underline{88.74}}$ &$\underset{\pm0.52}{32.52}$&$\underset{\pm1.70}{19.51}$&$\underset{\pm0.70}{24.55}$&$\underset{\pm0.73}{24.52}$  \\

         $\oplus\ T$ & $\underset{\pm0.52}{88.02}$& $\underset{\pm0.80}{\underline{35.15}}$& $\underset{\pm0.96}{\underline{25.17}}$& $\underset{\pm0.68}{\underline{29.96}}$& $\underset{\pm1.04}{\underline{31.49}}$\\ \midrule

        \textbf{ascc}$^*$ &$\underset{\pm0.36}{87.77}$ &$\underset{\pm0.64}{33.61}$&$\underset{\pm2.17}{15.13}$&$\underset{\pm0.77}{23.50}$&$\underset{\pm2.11}{26.80}$  \\

         $\oplus\ T$& $\underset{\pm0.80}{86.36}$& $\underset{\pm1.12}{\underline{34.93}}$& $\underset{\pm0.72}{\underline{27.36}}$& $\underset{\pm1.38}{\underline{30.93}}$& $\underset{\pm1.65}{\underline{33.46}}$\\ \midrule

        \textbf{dg-aug}$^*$ &$\underset{\pm0.39}{\underline{87.12}}$ &$\underset{\pm1.59}{34.74}$&$\underset{\pm1.83}{22.36}$&$\underset{\pm2.57}{26.11}$&$\underset{\pm0.75}{\underline{37.43}}$  \\

         $\oplus\ T$ & $\underset{\pm0.22}{87.09}$& $\underset{\pm2.64}{\underline{36.99}}$& $\underset{\pm2.86}{\underline{26.92}}$& $\underset{\pm1.67}{\underline{31.43}}$& $\underset{\pm1.90}{36.40}$\\ 

    \bottomrule
    \end{tabular}
    \caption{Adversarial Training (AT) combined with a training temperature of $T=200$ ($\oplus \ T$). For each adversarial attack, the higher adversarial accuracy between the AT model and the AT $\oplus \ T$ model is underlined. In almost all cases, the higher training temperature improves adversarial accuracy. Test-time calibration (\textit{cal}) is used to mitigate IOR. Dataset: Rotten Tomatoes.}
    \label{tab:AT-gradnorm-rt-app}
\end{table}

\begin{table}[t]
    \centering
    \small
    \begin{tabular}{l|c|cccc}
    \toprule
        Method & clean & bae & tf & pwws & dg \\ \midrule
        
        \textbf{baseline}$^*$ &  $\underset{\pm0.24}{\underline{93.13}}$ &$\underset{\pm0.85}{30.17}$&$\underset{\pm0.55}{5.77}$&$\underset{\pm2.01}{11.80}$&$\underset{\pm2.98}{8.32}$\\  
        $\oplus\ T$&$\underset{\pm0.89}{92.83}$ & $\underset{\pm0.95}{\underline{32.10}}$ & $\underset{\pm1.58}{\underline{6.42}}$ & $\underset{\pm1.20}{\underline{12.68}}$ & $\underset{\pm1.37}{\underline{8.45}}$  \\
 \midrule

        \textbf{pgd}$^*$& $\underset{\pm0.03}{\underline{93.48}}$ &$\underset{\pm0.43}{28.83}$&$\underset{\pm1.24}{4.88}$&$\underset{\pm0.69}{9.95}$&$\underset{\pm1.08}{5.45}$\\
        $\oplus\ T$ &$\underset{\pm0.10}{93.40}$ & $\underset{\pm0.65}{\underline{30.58}}$ & $\underset{\pm0.25}{\underline{5.43}}$ & $\underset{\pm0.99}{\underline{10.78}}$ & $\underset{\pm1.51}{\underline{6.33}}$  \\\midrule
        
        \textbf{freelb}$^*$ & $\underset{\pm0.23}{93.67}$ &$\underset{\pm1.00}{29.15}$&$\underset{\pm1.25}{4.93}$&$\underset{\pm0.30}{10.15}$&$\underset{\pm0.73}{\underline{5.48}}$\\
        $\oplus\ T$&$\underset{\pm0.10}{\underline{93.72}}$ & $\underset{\pm0.53}{\underline{30.23}}$ & $\underset{\pm0.03}{\underline{5.58}}$ & $\underset{\pm0.96}{\underline{10.78}}$ & $\underset{\pm0.98}{5.23}$  \\\midrule

        \textbf{ascc}$^*$ & $\underset{\pm0.57}{91.15}$ &$\underset{\pm0.23}{34.65}$&$\underset{\pm1.05}{4.60}$&$\underset{\pm0.22}{12.15}$&$\underset{\pm1.40}{11.28}$\\
        $\oplus\ T$&$\underset{\pm0.24}{\underline{91.78}}$ & $\underset{\pm0.03}{\underline{34.78}}$ & $\underset{\pm0.45}{\underline{7.57}}$ & $\underset{\pm0.64}{\underline{14.08}}$ & $\underset{\pm1.48}{\underline{11.55}}$  \\ \midrule

        \textbf{dg-aug}$^*$ &$\underset{\pm0.11}{\underline{92.58}}$ &$\underset{\pm2.82}{31.52}$&$\underset{\pm0.25}{4.68}$&$\underset{\pm0.11}{9.33}$&$\underset{\pm0.64}{\underline{29.45}}$  \\
        $\oplus\ T$&$\underset{\pm0.13}{91.98}$ & $\underset{\pm0.85}{\underline{31.88}}$ & $\underset{\pm0.28}{\underline{5.38}}$ & $\underset{\pm0.87}{\underline{9.40}}$ & $\underset{\pm1.26}{23.63}$  \\ 
        
    \bottomrule
    \end{tabular}
    \caption{Adversarial Training (AT) combined with a training temperature of $T=20$ ($\oplus \ T$). For each adversarial attack column the higher adversarial accuracy between the AT model and the AT $\oplus \ T$ model is underlined. In almost all cases, a higher training temperature improves adversarial accuracy. Test-time calibration (\textit{cal}) is used to mitigate IOR. Dataset: Emotion.}
    \label{tab:AT-gradnorm-twitter}
\end{table}

~\newpage
~\newpage
\newpage

\subsection{Transferability of High Temperature Training}
It is shown that training with a high temperature leads to a consistent gain in adversarial robustness to unseen adversarial attack forms. However, an adversary may attempt to exploit attack \textit{transferability} when looking to attack the target model trained with high temperature. To explore this notion of a transfer attack, with the \textit{rt} dataset, Table \ref{tab:transferability} shows the impact of finding adversarial examples for the source baseline ST$^*$ model and assessing their efficacy on the target baseline ST$^*$ $\oplus \ T$ model. It is evident from the significant increase in the adversarial accuracy for all the attack forms (\textit{bae},\textit{ tf}, \textit{pwws}, and \textit{dg}), that a transfer attack is not able to degrade the observed robustness gains for models trained with high temperatures.

\begin{table}[htb!]
    \centering
    \fontsize{8.5}{11}\selectfont
    \begin{tabular}{lc|cccc}
    \toprule
Source & Target  & bae & tf & pwws & dg\\ \midrule
baseline$^*$& baseline$^*$  & $\underset{\pm1.20}{31.39}$ & $\underset{\pm0.49}{17.82}$ & $\underset{\pm0.62}{20.42}$ & $\underset{\pm0.94}{20.11}$\\
$\oplus \ T$ & $\oplus \ T$  & $\underset{\pm0.84}{35.83}$ & $\underset{\pm4.57}{26.83}$ & $\underset{\pm3.07}{31.49}$ & $\underset{\pm4.71}{35.18}$\\ \midrule

baseline$^*$  & $\oplus \ T$  & $\underset{\pm0.30}{50.13}$ & $\underset{\pm0.38}{46.90}$ & $\underset{\pm1.22}{47.53}$ & $\underset{\pm1.13}{46.09}$\\
        \bottomrule
    \end{tabular}
    \caption{Transferability: adversarial examples are found for the \textit{source} model and evaluated on the \textit{target} model on the \textit{rt} test set. The results here demonstrate that the standard trained, high temperature ($\oplus T$) model's robustness gains relative to the baseline$^*$ model cannot be compromised by a transferability attack, i.e. the performance of the $\oplus T$ model are not degraded by adversarial examples generated from the baseline$^*$ model. Test-time calibration (\textit{cal}) is used to mitigate IOR.}
    \label{tab:heat-transferability}
\end{table}

\newpage

\subsection{Detailed Performance Breakdown} \label{app:heat-detailed-datasets}
Figure \ref{fig:temp} presents the adversarial accuracy of baseline ST$^*$ models trained with different training temperatures. In this section, for reference, we provide the detailed breakdown (average across 3 seeds and standard deviation) of performances for the different training temperatures for each dataset: \textit{rt} (Table \ref{tab:rt-pareto}), \textit{emotion} (Table \ref{tab:twitter-pareto}), \textit{cola} (Table \ref{tab:cola-pareto}), \textit{qnli} (Table \ref{tab:qnli-pareto}), and \textit{mrpc} (Table \ref{tab:mrpc-pareto}). These results are given for the DeBERTa model as in the main paper.

\begin{table}[htb!]
    \centering
    \small
    \begin{tabular}{l|c|cccc}
    \toprule
        Temp & clean & bae & tf & pwws & dg \\ \midrule
        1 & $\underset{\pm0.19}{88.56}$ & $\underset{\pm0.14}{32.40}$ & $\underset{\pm0.48}{18.79}$ & $\underset{\pm1.22}{21.36}$ & $\underset{\pm0.66}{21.11}$\\ % & $\underset{\pm0.20}{88.52}$ - morpheus
        10 & $\underset{\pm0.49}{88.18}$ & $\underset{\pm0.82}{34.12}$ & $\underset{\pm2.65}{23.23}$ & $\underset{\pm1.22}{24.71}$ & $\underset{\pm2.10}{26.52}$\\ % & $\underset{\pm0.49}{88.18}$ - morpheus
        20 & $\underset{\pm0.52}{88.46}$ & $\underset{\pm1.97}{34.24}$ & $\underset{\pm2.18}{23.14}$ & $\underset{\pm1.28}{26.70}$ & $\underset{\pm1.69}{29.49}$\\
        100 & $\underset{\pm0.70}{88.21}$ & $\underset{\pm0.81}{34.83}$ & $\underset{\pm3.79}{24.82}$ & $\underset{\pm2.86}{26.36}$ & $\underset{\pm4.36}{29.49}$\\
        200 & $\underset{\pm0.44}{87.55}$ & $\underset{\pm0.84}{35.83}$ & $\underset{\pm4.57}{26.83}$ & $\underset{\pm3.07}{31.49}$ & $\underset{\pm4.71}{35.18}$\\
        2000 & $\underset{\pm0.89}{86.40}$ &$\underset{\pm1.79}{35.46}$ &$\underset{\pm0.76}{25.02}$ &$\underset{\pm1.38}{29.99}$ &$\underset{\pm1.05}{30.81}$\\
        \bottomrule
    \end{tabular}
    \caption{\textbf{rt:} The use of a training temperature, $T$, is a simple adjustment in standard baseline model training (baseline$^*$), where the temperature parameter, $T$, is used to scale down predicted model logits. Higher training temperatures enhance model robustness against unseen adversarial attacks (\textit{bae, tf, pwws, dg}) without requiring prior knowledge of these attack forms during training. Results here report the clean and adversarial accuracy. Test-time calibration (\textit{cal}) is used to mitigate IOR.}
    \label{tab:rt-pareto}
\end{table}

\begin{table}[htb!]
    \centering
    \small
    \begin{tabular}{l|c|cccc}
    \toprule
        Temp & clean & bae & tf & pwws & dg \\ \midrule
        1 & $\underset{\pm0.10}{92.72}$ & $\underset{\pm0.20}{31.55}$ & $\underset{\pm1.30}{6.53}$ & $\underset{\pm1.31}{11.85}$ & $\underset{\pm1.22}{8.20}$\\
        2 & $\underset{\pm0.36}{92.72}$ & $\underset{\pm0.83}{31.33}$ & $\underset{\pm1.66}{6.45}$ & $\underset{\pm2.26}{12.60}$ & $\underset{\pm2.43}{8.95}$\\
        20 & $\underset{\pm0.89}{92.83}$ & $\underset{\pm0.95}{32.10}$ & $\underset{\pm1.58}{6.42}$ & $\underset{\pm1.20}{12.68}$ & $\underset{\pm1.37}{8.45}$\\
        100 & $\underset{\pm0.15}{90.98}$ & $\underset{\pm0.79}{31.50}$ & $\underset{\pm0.10}{7.55}$ & $\underset{\pm0.74}{14.43}$ & $\underset{\pm1.87}{9.23}$\\
        200 & $\underset{\pm0.24}{85.67}$ & $\underset{\pm0.47}{32.87}$ & $\underset{\pm0.64}{11.82}$ & $\underset{\pm0.66}{16.28}$ & $\underset{\pm1.44}{10.90}$\\
        \bottomrule
    \end{tabular}
    \caption{\textbf{emotion:} The use of a training temperature, $T$, is a simple adjustment in standard baseline model training (baseline$^*$), where the temperature parameter, $T$, is used to scale down predicted model logits. Higher training temperatures enhance model robustness against unseen adversarial attacks (\textit{bae, tf, pwws, dg}) without requiring prior knowledge of these attack forms during training. Results here report the clean and adversarial accuracy. Test-time calibration (\textit{cal}) is used to mitigate IOR.}
    \label{tab:twitter-pareto}
\end{table}

\begin{table}[htb!]
    \centering
    \small
    \begin{tabular}{l|c|cccc}
    \toprule
        Temp & clean & bae & tf & pwws & dg \\ \midrule
        1 & $\underset{\pm0.53}{83.70}$ & $\underset{\pm0.59}{3.39}$ & $\underset{\pm0.43}{5.43}$ & $\underset{\pm0.73}{10.23}$ & $\underset{\pm1.49}{11.63}$\\
        10 & $\underset{\pm0.72}{84.21}$ & $\underset{\pm0.83}{6.30}$ & $\underset{\pm0.48}{9.17}$ & $\underset{\pm0.42}{11.98}$ & $\underset{\pm1.75}{13.45}$\\
        100 & $\underset{\pm0.91}{82.68}$ & $\underset{\pm2.34}{7.41}$ & $\underset{\pm3.22}{9.04}$ & $\underset{\pm1.08}{11.89}$ & $\underset{\pm4.96}{13.04}$\\
        \bottomrule
    \end{tabular}
    \caption{\textbf{cola:} The use of a training temperature, $T$, is a simple adjustment in standard model training (baseline$^*$), where the temperature parameter, $T$, is used to scale down predicted model logits. Higher training temperatures enhance model robustness against unseen adversarial attacks (\textit{bae, tf, pwws, dg}) without requiring prior knowledge of these attack forms during training. Results here report the clean and adversarial accuracy. Test-time calibration (\textit{cal}) is used to mitigate IOR.} 
    \label{tab:cola-pareto}
\end{table}

\begin{table}[htb!]
    \centering
    \small
    \begin{tabular}{l|c|cccc}
    \toprule
        Temp & clean & bae & tf & pwws & dg \\ \midrule
        1 & $\underset{\pm0.26}{93.17}$ & $\underset{\pm1.88}{35.71}$ & $\underset{\pm3.17}{20.71}$ & $\underset{\pm2.38}{19.79}$ & $\underset{\pm4.39}{17.92}$\\
        10 & $\underset{\pm0.94}{92.92}$ & $\underset{\pm0.66}{35.00}$ & $\underset{\pm1.56}{18.92}$ & $\underset{\pm1.02}{17.38}$ & $\underset{\pm2.76}{16.38}$\\
        20 & $\underset{\pm0.66}{92.75}$ & $\underset{\pm0.38}{35.75}$ & $\underset{\pm1.54}{20.17}$ & $\underset{\pm0.90}{19.29}$ & $\underset{\pm1.19}{16.75}$\\
        50 & $\underset{\pm0.75}{93.00}$ & $\underset{\pm1.39}{36.67}$ & $\underset{\pm1.70}{23.46}$ & $\underset{\pm0.26}{22.54}$ & $\underset{\pm1.56}{19.79}$\\
        100 & $\underset{\pm0.33}{92.75}$ & $\underset{\pm1.19}{37.83}$ & $\underset{\pm0.38}{22.96}$ & $\underset{\pm1.44}{22.00}$ & $\underset{\pm1.11}{21.63}$\\
        150 & $\underset{\pm0.70}{92.38}$ & $\underset{\pm2.00}{35.54}$ & $\underset{\pm2.12}{22.08}$ & $\underset{\pm2.32}{19.71}$ & $\underset{\pm3.59}{18.29}$\\
        200 & $\underset{\pm0.47}{92.96}$ & $\underset{\pm1.21}{35.92}$ & $\underset{\pm3.54}{21.33}$ & $\underset{\pm2.48}{21.04}$ & $\underset{\pm3.06}{19.75}$\\
        \bottomrule
    \end{tabular}
    \caption{\textbf{qnli:} The use of a training temperature, $T$, is a simple adjustment in standard model training (baseline$^*$), where the temperature parameter, $T$, is used to scale down predicted model logits. Higher training temperatures enhance model robustness against unseen adversarial attacks (\textit{bae, tf, pwws, dg}) without requiring prior knowledge of these attack forms during training. Results here report the clean and adversarial accuracy. Test-time calibration (\textit{cal}) is used to mitigate IOR.}
    \label{tab:qnli-pareto}
\end{table}

\begin{table}[htb!]
    \centering
    \small
    \begin{tabular}{l|c|cccc}
    \toprule
        Temp & clean & bae & tf & pwws & dg \\ \midrule
        1  & $\underset{\pm0.26}{87.46}$& $\underset{\pm1.94}{46.42}$ & $\underset{\pm3.92}{38.83}$ & $\underset{\pm3.25}{28.63}$ & $\underset{\pm6.11}{33.50}$ \\
        5 &  $\underset{\pm0.44}{87.79}$ &  $\underset{\pm3.05}{46.17}$ &  $\underset{\pm4.58}{38.88}$ &  $\underset{\pm2.20}{30.42}$ &  $\underset{\pm1.11}{31.50}$\\
        10 & $\underset{\pm0.31}{88.21}$& $\underset{\pm3.62}{48.67}$ & $\underset{\pm5.63}{43.92}$ & $\underset{\pm3.69}{32.75}$ & $\underset{\pm5.03}{35.04}$  \\
        50 & $\underset{\pm0.29}{86.92}$ & $\underset{\pm4.15}{45.54}$ & $\underset{\pm8.15}{31.88}$ & $\underset{\pm5.00}{24.63}$ & $\underset{\pm7.22}{29.21}$\\
        100 & $\underset{\pm0.94}{86.21}$& $\underset{\pm2.58}{48.25}$ & $\underset{\pm6.39}{35.63}$ & $\underset{\pm5.59}{26.08}$ & $\underset{\pm4.74}{27.96}$  \\
        \bottomrule
    \end{tabular}
    \caption{\textbf{mrpc:} The use of a training temperature, $T$, is a simple adjustment in standard model training (baseline$^*$), where the temperature parameter, $T$, is used to scale down predicted model logits. Higher training temperatures enhance model robustness against unseen adversarial attacks (\textit{bae, tf, pwws, dg}) without requiring prior knowledge of these attack forms during training. Results here report the clean and adversarial accuracy. Test-time calibration (\textit{cal}) is used to mitigate IOR.}
    \label{tab:mrpc-pareto}
\end{table}

In Table \ref{tab:agnews-pareto}, we further include results on a 6th dataset AGNews~\citep{Zhang2015CharacterlevelCN}, where there are four news classes, 96k training samples, 24k validation samples and 7.6k test samples. For this dataset, it can be observed that a high training temperature is not a successful method unless a fraction of the dataset (10k training samples) is used during training. Future work is necessary to understand the nature of this specific dataset or other similar datasets that led to such a different behaviour for the temperature training approach.

\begin{table}[htb!]
    \centering
    \small
    \begin{tabular}{l|c|cccc}
    \toprule
        Temp & clean & bae & tf & pwws & dg \\ \midrule
        % 0.1 & NaN & - & - & - & - \\
        % 0.5 & NaN & - & - & - & - \\
        % 0.75 & $\underset{\pm0.31}{93.79}$ & $\underset{\pm0.57}{81.88}$ & $\underset{\pm1.56}{29.75}$ & $\underset{\pm0.82}{42.13}$ & $\underset{\pm1.69}{39.63}$\\ \midrule
        1 & $\underset{\pm0.22}{93.88}$ & $\underset{\pm0.25}{81.50}$ & $\underset{\pm0.19}{29.46}$ & $\underset{\pm2.19}{43.00}$ & $\underset{\pm2.89}{39.08}$\\ \midrule
        1.5 & $\underset{\pm0.13}{93.75}$ & $\underset{\pm0.51}{80.92}$ & $\underset{\pm3.26}{29.08}$ & $\underset{\pm5.20}{42.13}$ & $\underset{\pm3.19}{38.58}$\\
        2 & $\underset{\pm0.07}{93.92}$ & $\underset{\pm0.56}{80.04}$ & $\underset{\pm3.80}{25.00}$ & $\underset{\pm3.19}{40.38}$ & $\underset{\pm5.69}{38.54}$\\
        20 & $\underset{\pm0.36}{93.83}$ & $\underset{\pm1.02}{79.25}$ & $\underset{\pm2.07}{23.50}$ & $\underset{\pm2.60}{37.58}$ & $\underset{\pm4.56}{34.58}$\\
        \bottomrule
    \end{tabular}
    \caption{\textbf{agnews:} The use of a training temperature, $T$, is a simple adjustment in standard model training (baseline$^*$), where the temperature parameter, $T$, is used to scale down predicted model logits. Higher training temperatures enhance model robustness against unseen adversarial attacks (\textit{bae, tf, pwws, dg}) without requiring prior knowledge of these attack forms during training. Results here report the clean and adversarial accuracy. Test-time calibration (\textit{cal}) is used to mitigate IOR.}
    \label{tab:agnews-pareto}
\end{table}

% \begin{table}[htb!]
%     \centering
%     \begin{tabular}{l|c|cccc}
%     \toprule
%         Temp & clean & bae & tf & pwws & dg \\ \midrule
%         1 & $\underset{\pm0.07}{93.33}$ & $\underset{\pm1.09}{81.46}$ & $\underset{\pm3.81}{26.25}$ & $\underset{\pm3.25}{37.67}$ & $\underset{\pm3.18}{31.08}$\\
%         2 & $\underset{\pm0.29}{93.58}$ & $\underset{\pm0.33}{81.50}$ & $\underset{\pm3.89}{24.54}$ & $\underset{\pm2.17}{37.89}$ & $\underset{\pm1.02}{30.83}$\\
%         20 & $\underset{\pm0.22}{93.13}$ & $\underset{\pm1.30}{78.46}$ & $\underset{\pm2.07}{21.13}$ & $\underset{\pm0.80}{31.67}$ & $\underset{\pm3.06}{29.38}$\\
%         \bottomrule
%     \end{tabular}
%     \caption{\textbf{AGNews:} Impact of training temperature. Note, gradient normalization is always used during training and the models are calibrated at test-time to ensure no risk of IOR. \textbf{Eval Epoch 3}}
%     \label{tab:agnews-3}
% \end{table}

\begin{table}[htb!]
    \centering
    \small
    \begin{tabular}{l|c|cccc}
    \toprule
        Temp & clean & bae & tf & pwws & dg \\ \midrule
        1 & $\underset{\pm0.38}{93.17}$ & $\underset{\pm0.54}{78.00}$ & $\underset{\pm2.32}{32.33}$ & $\underset{\pm0.47}{42.08}$ & $\underset{\pm2.53}{40.54}$\\
        % 2 & $\underset{\pm0.40}{92.83}$ & $\underset{\pm0.44}{78.17}$ & $\underset{\pm2.90}{30.83}$ & $\underset{\pm5.41}{38.92}$ & $\underset{\pm2.43}{38.29}$ \\
        10 & $\underset{\pm0.40}{92.08}$ & $\underset{\pm0.66}{79.00}$ & $\underset{\pm2.89}{38.33}$& $\underset{\pm2.09}{50.42}$ & $\underset{\pm0.63}{46.54}$\\
        20 & $\underset{\pm0.19}{92.46}$ & $\underset{\pm0.69}{77.92}$ & $\underset{\pm2.63}{38.33}$ & $\underset{\pm2.89}{49.21}$ & $\underset{\pm1.61}{45.67}$ \\
        100 & $\underset{\pm0.38}{92.13}$ & $\underset{\pm0.00}{77.50}$ & $\underset{\pm2.81}{30.33}$& $\underset{\pm0.76}{41.46}$ & $\underset{\pm2.34}{40.38}$\\
        \bottomrule
    \end{tabular}
    \caption{\textbf{agnews:} The use of a training temperature, $T$, is a simple adjustment in standard model training (ST$^*$), where the temperature parameter, $T$, is used to scale down predicted model logits. Higher training temperatures enhance model robustness against unseen adversarial attacks (\textit{bae, tf, pwws, dg}) without requiring prior knowledge of these attack forms during training. Results here report the clean and adversarial accuracy. Training with 10k samples - 1/10th of default agnews training set size. Test-time calibration (\textit{cal}) is used to mitigate IOR.}
    \label{tab:agnews-sample}
\end{table}

\newpage

\subsection{Reproducing with Other Models} \label{app:heat-models}

The main paper presents results using the DeBERTa model. Here we repeat the core experiments on other popular \textit{baseline} models: BERT (Table \ref{tab:bert}) and RoBERTa (Table \ref{tab:roberta}). The results here are presented for the \textit{rt} dataset.

\begin{table}[htb!]
    \centering
    \small
    \begin{tabular}{l|c|cccc}
    \toprule
        Temp & clean & bae & tf & pwws & dg \\ \midrule
        1 & $\underset{\pm0.50}{85.08}$ & $\underset{\pm0.76}{30.52}$ & $\underset{\pm0.32}{21.01}$ & $\underset{\pm0.34}{21.20}$ & $\underset{\pm2.14}{23.14}$\\
        10 & $\underset{\pm0.58}{84.79}$ & $\underset{\pm0.66}{32.16}$ & $\underset{\pm1.23}{25.88}$ & $\underset{\pm1.89}{23.96}$ & $\underset{\pm1.67}{27.48}$\\
        100 & $\underset{\pm0.54}{84.76}$ & $\underset{\pm0.78}{33.01}$ & $\underset{\pm1.99}{27.12}$ & $\underset{\pm2.45}{25.88}$ & $\underset{\pm2.02}{28.92}$\\
        \bottomrule
    \end{tabular}
    \caption{\textbf{BERT:} The use of a training temperature, $T$, is a simple adjustment in standard model training (ST$^*$), where the temperature parameter, $T$, is used to scale down predicted model logits. Higher training temperatures enhance model robustness against unseen adversarial attacks (\textit{bae, tf, pwws, dg}) without requiring prior knowledge of these attack forms during training. Results here report the clean and adversarial accuracy. Test-time calibration (\textit{cal}) is used to mitigate IOR. Result for \textit{rt} dataset.}
    \label{tab:bert}
\end{table}

\begin{table}[htb!]
    \centering
    \small
    \begin{tabular}{l|c|cccc}
    \toprule
        Temp & clean & bae & tf & pwws & dg \\ \midrule
        1 & $\underset{\pm0.47}{88.27}$ & $\underset{\pm0.74}{32.46}$ & $\underset{\pm0.72}{17.01}$ & $\underset{\pm0.05}{21.23}$ & $\underset{\pm1.71}{24.30}$\\
        10 & $\underset{\pm0.65}{88.25}$ & $\underset{\pm0.86}{33.17}$ & $\underset{\pm1.86}{21.96}$ & $\underset{\pm1.15}{24.32}$ & $\underset{\pm3.02}{28.85}$\\
        100 & $\underset{\pm0.72}{88.26}$ & $\underset{\pm0.92}{33.55}$ & $\underset{\pm2.04}{23.20}$ & $\underset{\pm2.12}{26.03}$ & $\underset{\pm3.55}{29.66}$\\
        \bottomrule
    \end{tabular}
    \caption{\textbf{RoBERTa:} The use of a training temperature, $T$, is a simple adjustment in standard model training (ST$^*$), where the temperature parameter, $T$, is used to scale down predicted model logits. Higher training temperatures enhance model robustness against unseen adversarial attacks (\textit{bae, tf, pwws, dg}) without requiring prior knowledge of these attack forms during training. Results here report the clean and adversarial accuracy. Test-time calibration (\textit{cal}) is used to mitigate IOR. Result for \textit{rt} dataset.}
    \label{tab:roberta}
\end{table}

\newpage

\subsection{High Temperature Training GradNorm Ablation} \label{sec:heat-gradnorm-ablation}

We find that including gradient normalization during training with a high temperature is beneficial in preventing a decrease in clean accuracy, whilst maintaining the gains in genuine adversarial robustness. This is demonstrated in the results in Table \ref{tab:heat-gradnorm-ablation}, where calibration is used at test-time to ensure there is no IOR.

\begin{table}[htb!]
    \centering
    \small
    \begin{tabular}{lccccc}
    \toprule
        &  clean & bae & tf & pwws & dg\\ \midrule
       baseline $\oplus T$  & $\underset{\pm0.20}{85.12}$ & $\underset{\pm1.77}{35.71}$ & $\underset{\pm5.72}{26.85}$ & $\underset{\pm5.14}{31.26}$ & $\underset{\pm6.20}{35.82}$\\
       baseline$^*$ $\oplus T$ & $\underset{\pm0.44}{87.55}$ & $\underset{\pm0.84}{35.83}$ & $\underset{\pm4.57}{26.83}$ & $\underset{\pm3.07}{31.49}$ & $\underset{\pm4.71}{35.18}$\\
       \bottomrule
    \end{tabular}
    \caption{Gradnorm ($^*$) Ablation. Results are presented for the baseline system for the \textit{rt} dataset and the DeBERTa model. A high training temperature of $T=200$ is used. Test-time calibration (\textit{cal}) is used to mitigate IOR.}
    \label{tab:heat-gradnorm-ablation}
\end{table}

\subsection{IOR From High Temperature Training} \label{sec:heat-cal}

In the main paper we present high temperature training as an effective method to induce genuine adversarial robustness. Here we demonstrate the need to apply the test-time calibration methods of Section \ref{sec:mitigate} to ensure there is no IOR. From Table \ref{tab:heat-cal}, we see an approximately two-fold increase in adversarial accuracy when no calibration is applied before evaluation.

\begin{table}[h]
    \centering
    \small
    \fontsize{7}{11}\selectfont
    \begin{tabular}{lc|c|cccc}
    \toprule
       \textbf{Method} & \textbf{Adv.} & \textbf{clean} & \textbf{bae} & \textbf{tf} & \textbf{pwws} & \textbf{dg} \\ \midrule

        baseline$^*$ $\oplus$T &-& $\underset{\pm0.44}{87.55}$ & $\underset{\pm0.81}{38.87}$ & $\underset{\pm0.41}{56.62}$ & $\underset{\pm0.49}{61.01}$ & $\underset{\pm1.41}{67.20}$\\
        
        &cal& $\underset{\pm0.44}{87.55}$ & $\underset{\pm0.84}{35.83}$ & $\underset{\pm4.57}{26.83}$ & $\underset{\pm3.07}{31.49}$ & $\underset{\pm4.71}{35.18}$\\
        \bottomrule
    \end{tabular}
    \caption{Adversarial Accuracy of high training temperature method ($T=200$) with test-time calibration: none (-) or temperature scaling (-). \textbf{rt} dataset on DeBERTa model.}
    \label{tab:heat-cal}
\end{table}

\newpage

\subsection{Other Ablations}

Augmentation based adversarial training approaches, such as dg-aug$^*$ in the main paper, have twice as many training steps (due to there being double the training set size). To match the standard training setting, in Table \ref{tab:AT-gradnorm-halfsteps} we evaluated the training with high temperature approach combined with dg-aug$^*$ at half the number of training steps. Similarly, in Table \ref{tab:base-double} we consider the inverse setting, where we double the number of training iterations for the \textit{baseline}$^*$ model (in standard training), as well as linearly scaling the learning rate scheduler across the increased number of iterations.

\begin{table}[htb!]
    \centering
    \small
    \begin{tabular}{lc|c|cc}
    \toprule
        Method & iters & clean & pwws & dg \\ \midrule
        
        baseline$^*$ $\oplus T$  & default & $\underset{\pm0.44}{87.55}$ & $\underset{\pm3.07}{31.49}$ & $\underset{\pm4.71}{35.18}$ \\ \midrule
        
        dg-aug$^*$ $\oplus T$ & default & $\underset{\pm0.22}{87.09}$ & $\underset{\pm1.67}{31.43}$ & $\underset{\pm1.90}{36.40}$\\
         & half & $\underset{\pm0.44}{86.05}$ & $\underset{\pm5.21}{37.02}$ & $\underset{\pm2.36}{43.00}$\\
         
        % pwws-aug(1) +gradnorm $T^{(tr)}=200$ & default & $\underset{\pm0.34}{86.02}$ & $\underset{\pm0.81}{40.84}$ & $\underset{\pm0.62}{39.81}$\\
        %  & half & $\underset{\pm0.11}{85.24}$ & $\underset{\pm7.47}{37.68}$ & $\underset{\pm6.69}{38.49}$\\

    \bottomrule
    \end{tabular}
    \caption{Matched number of iterations for baseline$^*$ and high temperature training with dg-aug$^*$ by halving the number of training steps for dg-aug$^*$. Test-time calibration (\textit{cal}) is used to mitigate IOR.}
    \label{tab:AT-gradnorm-halfsteps}
\end{table}

\begin{table}[htb!]
    \centering
    \fontsize{7}{11}\selectfont
    \begin{tabular}{lc|c|cccc}
    \toprule
        Method & Epochs & clean &bae&tf& pwws & dg \\ \midrule
        \textbf{baseline}$^*$& 5 & $\underset{\pm0.19}{88.56}$ & $\underset{\pm0.14}{32.40}$ & $\underset{\pm0.48}{18.79}$ & $\underset{\pm1.22}{21.36}$ & $\underset{\pm0.66}{21.11}$\\
        & 10 & $\underset{\pm0.62}{88.34}$ & $\underset{\pm0.52}{33.61}$ & $\underset{\pm0.50}{18.76}$ & $\underset{\pm0.61}{22.39}$ & $\underset{\pm0.86}{23.45}$\\ \cmidrule{2-7}
        $\oplus T$  & 5  & $\underset{\pm0.44}{87.55}$ & $\underset{\pm0.84}{35.83}$ & $\underset{\pm4.57}{26.83}$ & $\underset{\pm3.07}{31.49}$ & $\underset{\pm4.71}{35.18}$\\
         & 10 & $\underset{\pm0.33}{87.55}$ & $\underset{\pm1.31}{34.43}$ & $\underset{\pm2.16}{25.48}$ & $\underset{\pm2.19}{30.11}$ & $\underset{\pm4.60}{33.40}$\\ \midrule
         
        \textbf{dg-aug}$^*$ & - &$\underset{\pm0.39}{87.12}$ &$\underset{\pm1.59}{34.74}$&$\underset{\pm1.83}{22.36}$&$\underset{\pm2.57}{26.11}$&$\underset{\pm0.75}{37.43}$  \\
        $\oplus T$  & -& $\underset{\pm0.22}{87.09}$& $\underset{\pm2.64}{36.99}$& $\underset{\pm2.86}{26.92}$& $\underset{\pm1.67}{31.43}$& $\underset{\pm1.90}{36.40}$\\
    \bottomrule
    \end{tabular}
    \caption{Doubling training iterations for the baseline$^*$ model with scaled scheduler decay to match number of iterations in augmentation based AT. Test-time calibration (\textit{cal}) is used to mitigate IOR.}
    \label{tab:base-double}
\end{table}

\onecolumn
\section{Algorithms}

\subsection{Method for Temperature Scaling Optimization to Mitigate IOR} \label{app:opt}

\tiny
\begin{lstlisting}
        def optimize_temp(self, factor=10):
        '''return temperature optimized to be successful in dg attack'''

        adv_temp = 1
        acc = self.eval_attack(adv_temp)    

        # check whether we need to increase or decrease temp
        test_temp_acc = self.eval_attack(adv_temp*factor)
        if test_temp_acc < acc:
            left = adv_temp
            right = 1e6 # assumes this as a maximum
        else:
            left = 1e-10 # 0 causes floating point errors
            right = adv_temp

        # seach for optimal temp (minima adv acc) using Brent algorithm (faster convergence of golden section algorithm)
        opt_temp = scipy.optimize.brent(self.eval_attack, brack=(left, 0.5*(left+right), right), maxiter=10)

        return opt_temp
\end{lstlisting}

\subsection{Base Class Definition with High Temperature Training for Genuine Robustness}\label{app:code}

\tiny
\begin{lstlisting}

class BaseClassifier(nn.Module):
    def __init__(self, model_name='bert-base-uncased', num_labels=2, pretrained=True, temperature=1):
        super().__init__()
        self.model_name = model_name
        self.temperature = temperature
        if pretrained:
            self.model = AutoModelForSequenceClassification.from_pretrained(model_name, num_labels=num_labels)
            self.tokenizer = AutoTokenizer.from_pretrained(model_name)
        else:
            config = AutoConfig.from_pretrained(model_name, num_labels=num_labels) # returns config and not pretrained weights 
            self.model = AutoModelForSequenceClassification.from_config(config)
            self.tokenizer = AutoTokenizer.from_pretrained(model_name)
        self.config = AutoConfig.from_pretrained(model_name, num_labels=num_labels)
    
    def forward(self, input_ids=None, attention_mask=None, inputs_embeds=None):
        logits = self.model(input_ids, attention_mask=attention_mask, inputs_embeds=inputs_embeds)[0]
        logits = logits / self.temperature
        return logits


\end{lstlisting}

\end{document}